\documentclass{article}
\usepackage{microtype}
\usepackage{graphicx}
\usepackage{subfigure}
\usepackage{booktabs}

\usepackage{hyperref}

\usepackage{amsmath,amsfonts,bm}

\def\eqref#1{equation~\ref{#1}}
\def\ceil#1{\lceil #1 \rceil}

\def\1{\bm{1}}

\def\eps{{\epsilon}}

\def\vf{{\bm{f}}}
\def\vg{{\bm{g}}}

\def\vk{{\bm{k}}}

\def\vt{{\bm{t}}}

\def\vv{{\bm{v}}}

\def\vx{{\bm{x}}}

\def\mA{{\bm{A}}}
\def\mB{{\bm{B}}}

\def\mD{{\bm{D}}}

\def\mF{{\bm{F}}}
\def\mG{{\bm{G}}}

\def\mI{{\bm{I}}}
\def\mJ{{\bm{J}}}
\def\mK{{\bm{K}}}

\def\mS{{\bm{S}}}
\def\mT{{\bm{T}}}

\def\mX{{\bm{X}}}

\DeclareMathAlphabet{\mathsfit}{\encodingdefault}{\sfdefault}{m}{sl}
\SetMathAlphabet{\mathsfit}{bold}{\encodingdefault}{\sfdefault}{bx}{n}

\newcommand{\R}{\mathbb{R}}

\DeclareMathOperator*{\argmax}{arg\,max}

\DeclareMathOperator{\Tr}{Tr}

\usepackage{hyperref}
\usepackage{url}
\usepackage[acronym]{glossaries}

\newacronym{kpca}{Kernel PCA}{Kernel Principal Component Analysis}
\newacronym{cca}{CCA}{Canonical Component Analysis}
\newacronym{rkhs}{RKHS}{Reproducing Kernel Hilbert Space}
\newacronym{ntk}{NTK}{Neural Tangent Kernel}
\newacronym{fim}{FIM}{Fisher Information Matrix}
\newacronym{frnorm}{FR norm}{Fisher-Rao norm}
\newacronym{cka}{CKA}{Centred Kernel Alignment}
\newacronym{nbs}{NBS}{Normalised Bures Similarity}
\newacronym{psd}{PSD}{positive semi-definite}
\newacronym{cwt}{CWT}{Clarkson-Woodruff Transformation}
\newacronym{scka}{SCKA}{Sketched CKA}
\newacronym{JLT}{JLT}{Johnson-Lindenstrauss Transform}

\usepackage{tabularx}
\usepackage{amsmath,amssymb,amsthm}
\usepackage{bbm}
\usepackage{graphicx}
\usepackage{listings}
\usepackage{upquote}
\usepackage{array}
\usepackage{multicol}
\usepackage{multirow}
\usepackage{url}
\usepackage{color}
\usepackage{wrapfig,lipsum,booktabs}
\usepackage{makecell}
\usepackage{changepage}
\usepackage{xcolor}

\usepackage{tabulary}
\newcolumntype{K}[1]{>{\centering\arraybackslash}p{#1}}

\usepackage{tabu}
\usepackage{arydshln}

\usepackage[export]{adjustbox}

\usepackage[accepted]{icml2020}

\icmltitlerunning{Similarity of Neural Networks with Gradients}

\begin{document}

\twocolumn[
\icmltitle{Similarity of Neural Networks with Gradients}

\icmlsetsymbol{equal}{*}

\begin{icmlauthorlist}
\icmlauthor{Shuai Tang}{ucsd,equal}
\icmlauthor{Wesley J. Maddox}{nyu,equal}
\icmlauthor{Charlie Dickens}{warwick,equal}
\icmlauthor{Tom Diethe}{amzn}
\icmlauthor{Andreas Damianou}{amzn}
\end{icmlauthorlist}

\icmlaffiliation{ucsd}{Department of Cognitive Science, UC San Diego, La Jolla, CA, USA}
\icmlaffiliation{nyu}{Center for Data Science, New York University, New York, NY, USA}
\icmlaffiliation{warwick}{Department of Computer Science, University of Warwick, Coventry, England, UK}
\icmlaffiliation{amzn}{Amazon, Cambridge, England, UK}

\icmlcorrespondingauthor{Shuai Tang}{shuaitang93@ucsd.edu}

\icmlkeywords{neural networks, similarity index, kernel mean embedding}

\vskip 0.3in
]

\printAffiliationsAndNotice{\icmlEqualContribution}

\begin{abstract}
A suitable similarity index for comparing learnt neural networks plays an important role in understanding the behaviour of the highly-nonlinear functions, and can provide insights on further theoretical analysis and empirical studies. We define two key steps when comparing models: firstly, the \textit{representation} abstracted from the learnt model, where we propose to leverage both feature vectors and gradient ones (which are largely ignored in prior work) into designing the representation of a neural network. Secondly, we define the employed \textit{similarity index} which gives desired invariance properties, and we facilitate the chosen ones with sketching techniques for comparing various datasets efficiently. Empirically, we show that the proposed approach provides a state-of-the-art method for computing similarity of neural networks that are trained independently on different datasets and the tasks defined by the datasets.

\end{abstract}

\section{Introduction}
To understand and interpret the applicability of neural networks in various domains, or the transferability of a learnt neural network to a new problem, we want to compare the various types of functions
induced by trained neural networks.
In particular, we would like to compare different models trained on the same dataset or, reversely, compare models with the same architecture trained on different datasets.
The problem of unveiling the similarity between models decomposes in having to specify a suitable \textit{similarity index} and an informative \textit{representation} of a learnt neural network. A well-designed method for similarity comparison can be applied to numerous paradigms,
such as transfer learning \cite{Dwivedi2019RepresentationSA}, meta learning \cite{Achille2019Task2VecTE}, and imitation learning \cite{Osa2018AnAP}.

Given the above, we aim to develop a method for comparing neural networks that is composed of the following two key steps. Firstly, it forms a meaningful representation of a given model conditioned on a given dataset; this leads to the use of kernel matrices for learning compact representations. Secondly, our method compares these kernel matrices by using the well-studied approaches of \acrfull{cka} \cite{Cortes2012AlgorithmsFL} and \acrfull{nbs} \cite{Muzellec2018GeneralizingPE}. As will be discussed later on, in our work these approaches are augmented with so-called \emph{sketching} techniques which offer scalability and desired invariances (such as invariance to varying number of samples across datasets).
For our purpose, \emph{sketches} are a tool for dimensionality reduction by random linear
transforms of the input which approximately preserve properties of the input.
We choose this approach because sketches can be applied quickly, in particular,
while streaming through the input and yield accurate summaries of the input.
Overall, the proposed method is capable of measuring the similarity among models given a dataset or, among datasets given models trained on a dataset.

In more detail, the \emph{representation} of the network comes from a kernel function evaluated at given neural network features, resulting in a kernel matrix. Now the question becomes, what type of representation do we use for the network features themselves? We choose to combine two types of features: firstly, learnt feature vectors produced at a given layer of the pre-trained model and, secondly, the gradient vectors of the loss function w.r.t. the feature vectors. The two chosen vectors of a learnt model reflect both the training data (feature vectors) and the learning task defined by the dataset where the given model is trained on (gradient vectors). To leverage both kinds of information for a single \textit{representation}, a simple starting point is to construct two separate kernel matrices for the feature and gradient vectors respectively, and then merge them using the Hadamard product.

The \textit{similarity index} is applied on the representations of different networks, as defined above, to quantify the corresponding similarity. Our similarity index method is based on the previously proposed \acrshort{cka} or \acrshort{nbs}. However, those previously proposed indices have an obvious drawback: they cannot handle comparisons where the undelying datasets are of different sizes. We tackle this issue by augmenting the indices with a family of widely-used sketching techniques.

Specifically, a sketch allows us to project two representations we wish to compare
to a chosen dimension {\small $M$}.
Typically, we can set {\small $M \ll N$} which results in a summary of the data
which is significantly smaller and thus, expensive matrix operations become
more tractable.
Crucially, we will use sketches which are fast to apply to the data, and a
projection dimension {\small $M$} which (approximately) preserves spectral structure of
the larger matrix.

Apart from invariance to \textbf{varying number of samples}, solved with sketching as discussed above, we also care about invariance to any \textbf{rotation} and \textbf{permutation} of neural network's hidden units, since these give equivalent function representations. Both \acrshort{cka} and \acrshort{nbs} with a linear kernel as similarity indices result in invariance to \textbf{isotropic scaling}, which fulfills our invariance requirements and justifies our choice of index. Therefore, our proposed method can be applied to compare different models trained on different datasets since it is capable of dealing with various situations.

Empirically, we observe that our proposed method provides meaningful similarity scores in terms of illustrating relationship among models based on both the training datasets and the learning tasks.
Additional analysis on the induced mapping from the designed kernel matrix provides insights for understanding our proposed methods.
Our code is available at: \href{https://github.com/amzn/xfer/tree/master/nn_similarity_index}{https://github.com/amzn/xfer/tree/master/nn\_similarity\_index}.

\section{Related Work}
Similarity of neural networks has been studied from different perspectives, including \acrfull{kpca}
of feature vectors produced from neural networks and functional analysis on the mapping implemented by them. \citet{Raghu2017SVCCASV} and \citet{Morcos2018InsightsOR} have focussed on computing correlation scores based on the feature representations generated by neural networks, and kernel-based methods are proposed to project the learnt feature vectors to a \acrfull{rkhs} where similarity can be better expressed \cite{Gretton2005MeasuringSD} depending on the chosen kernel.

The training of neural networks is commonly done via gradient descent, and the vector field of the Jacobian matrix of the neural network w.r.t. the parameters {\small $\mJ_\theta$} describes the loss landscape. This can be used for gradient-based kernels, including the Fisher kernel \cite{Jaakkola1998ExploitingGM} and \acrfull{ntk} \cite{Jacot2018NeuralTK}. Applications on large-scale learning tasks \cite{perronnin2010improving} demonstrate the effectiveness of such gradient information; however, it has not been widely applied to the comparison of neural networks due to its computational complexity.

A recent effort comes from
\citet{Achille2019Task2VecTE} where the diagonal terms in the Fisher Information matrix (\acrshort{fim}) w.r.t. the parameters in a reference model on individual tasks constitute the representations (Task2Vec), and cosine similarity is chosen as the similarity index. The \acrshort{fim} is defined as:
\begin{align}
 \mathbb{E}_{p(\vx)p_\theta(y|\vx)}\left[\nabla_\theta\log p_\theta(y|\vx)\nabla_\theta \log p_\theta(y|\vx)^\top\right],
\end{align}
and the vector representation {\small $\vv$} defined in Task2Vec of a model is computed as:
\begin{align}
 &\vv = [\vv_1, \vv_2, ..., \vv_{|\theta|}]^\top, \\
 &\text{where } \vv_i = \mathbb{E}_{p(\vx)p_{\theta_i}(y|\vx)} | \nabla_{\theta_i}\log p_{\theta_i}(y|\vx) |^2, \nonumber
\end{align}
where {\small $|\theta|$} is the number of parameters.
 The authors showed the effectiveness of the Task2Vec method on large-scale meta learning, and demonstrate the hierarchy of relavant vision tasks. The crucial drawback is that a reference model is required, and it fails to give reasonable similarity scores when multiple models are provided. Moreover, in contrast to our approach, Taks2Vec cannot be easily parallelised, since the required gradient is computed w.r.t. the parameters.

Our proposed representation for a learnt model on a chosen dataset leverages both the feature vectors and the gradient w.r.t. them in an efficient manner. \citet{pmlr-v89-liang19a} showed that the proposed \acrfull{frnorm}, which combines the information from both the features and gradients, is an indicator of the complexity and the generalisation ability of a model, and it is defined as:
\begin{align}
  ||\theta||_\text{fr}^2 = (L+1)^2\mathbb{E}\langle f_\theta(\vx), \nabla_{f_\theta(\vx)} \ell(f_\theta(\vx), y)\rangle^2,
\end{align}
where {\small $L$} is the number of layers in a given neural network.
\acrshort{frnorm} is invariant to re-parametrisation of models as long as the new parameterisation implements exactly the same function.  Under mild conditions on the activation functions, neural networks can also be studied through the \acrshort{frnorm}.
 The concern is that, if two models have exactly the same \acrshort{frnorm}, it doesn't necessarily mean that they are implementing the exact same function. A simple example is that a deep \emph{linear} neural network has the same \acrshort{frnorm} across all layers, however, the implemented function is varying as more layers are included.
Therefore, for comparing models, a scalar-representation as such is not sufficient.

The existence of prior work and the importance of the gradient vectors convince us that the representation abstracted from a neural network should also include the gradient information. In our proposed method, instead of only considering feature vectors as the representation of a neural network, we construct a kernel matrix depending on both the training data (feature) and the learning task (gradient), and we empirically show that it is considerably more appropriate in terms of comparing models.

\citet{Kornblith2019SimilarityON} evaluated several similarity indices primarily used in comparing feature maps learnt in a neural network model on the same dataset. Their results show that \acrshort{cka} \citep{Cortes2012AlgorithmsFL} is better at conveying the relations between feature maps than conventional methods including linear \acrfull{cca} and Kernel \acrshort{cca}. However, \acrshort{cka} is limited to comparing models trained on the same dataset.
Therefore, we aim to study the plausibility of adapting \acrshort{cka}
by sketching two different datasets of size $N_1, N_2$ into two matrices
of the same size, $M$, and then computing similarity scores using
the \acrshort{cka}.

\section{Method}
The proposed comparison can be directly applied to compute the similarity between any two deep learning models, given both on the dataset and on the task. Thus, our index has two realistic use cases: 1) understanding the similarity between deep learning models and 2)
understanding the difficulty of finetuning a pretrained model to another dataset.

\subsection{Representation}
We first introduce the \emph{representation} of a trained neural network given a dataset.

\textbf{Features and Gradients.} Provided a neural network $\phi$ pre-trained on a dataset {\small $D=\{(\vx_n, y_n)\}_{n=1}^N$},
in contrast to many previously proposed similarity indices which focus on comparing learnt feature maps generated from intermediate layers of neural networks,
our proposed method leverages both the feature map at a certain layer, and the gradient of the loss function w.r.t. the output of the chosen layer:
\begin{align}
  (\vf_l)_n &= \phi_l(\vx_n), \\
  (\vg_l)_n &= \mathbb{E}_{q_\theta(y|\vx_n)}\nabla_{(\vf_l)_n}\ell(\phi_\theta(\vx_n), y)
\end{align}
where {\small $(\vf_l)_n$} is the feature vector of the input {\small $\vx_n$} produced by a pre-trained network {\small $\phi$} parametrised with {\small $\theta$} at a $l$-th layer, {\small $(\vg_l)_n$} is the gradient. Both {\small $(\vf_l)_n, (\vg_l)_n \in \mathbb{R}^{d_l\times 1}$}, and {\small $d_l$} is the dimension of the output of $l$-th layer.

A special care needs to be taken since, when the loss function {\small $\ell(\phi_\theta(\vx_n), y))$} is the log-likelihood {\small $\log p_\theta(y|\vx_n)$}, the gradient tends towards zero: {\small $\mathbb{E}_{p_\theta(y|\vx)}\nabla_\vx \log p_\theta(y|\vx)=0$}, and {\small $(\vg_l)_n$} becomes useless.
Therefore, we pick
{\small $q_\theta(y|\vx_n)=p_\theta(y|\vx_n)^\beta/\int_Y p_\theta(y|\vx_n)^\beta dy$} for continuous outputs and {\small $q_\theta(y|\vx_n)=p_\theta(y|\vx_n)^\beta/\sum_{y=1}^C p_\theta(y|\vx_n)^\beta$} for categorical outputs where $C$ is the total number of categories. When {\small $\beta\in (0,1)$}, it gives a smoothed predictive distribution, and we set {\small $\beta=0.5$} in all our experiments.

Note that, here, {\small $(\vg_l)_n$} is not the Jacobian matrix {\small $\mJ_\theta\in\mathbb{R}^{N\times |\theta|}$} used in \acrfull{fim} \cite{edgeworth1908probable} or \acrfull{ntk} \cite{Jacot2018NeuralTK},
where {\small $|\theta|$} is the number of parameters;
{\small $(\vg_l)_n$} is computed w.r.t. the feature vectors and not the parameters.
Practically speaking, evaluating {\small $(\vg_l)_n$} only requires a single backward computation in most deep learning frameworks, whilst {\small $\mJ_\theta$} could take {\small $N$} backward calls which is computationally expensive.

An intuitive explanation of these two sources of information described in our method is that, {\small $(\vf_l)_n$} encodes the locations of the projected samples in the feature space, and {\small $(\vg_l)_n$} gives the direction and the intensity of where each projected sample should move to minimise the loss. Since the following description applies to every layer, we omit the subscription $l$ for simplicity.

\textbf{Kernel matrices.} Consider two kernels {\small $k_f(\cdot, \cdot)$} and {\small $k_g(\cdot, \cdot)$} which induce two mappings {\small $\vx\rightarrow \phi_f(\vx)$} and {\small $\vx\rightarrow \phi_g(\vx)$}, respectively.
In our case, we define two kernel matrices {\small $\mK_f, \mK_g\in\mathbb{R}^{N\times N}$} evaluated on
$D$ as follows:
\begin{align}
  \mK_f = \Phi_f^\top\Phi_f\succeq 0 \text{ and } \mK_g = \Phi_g^\top\Phi_g \succeq 0,
\end{align}
where { $\Phi_f=[\vf_1,\vf_2,...,\vf_N]\in\mathbb{R}^{d\times N}$} is the mapping produced by the learnt neural network and {\small $\Phi_g=[\vg_1,\vg_2,...,\vg_N]\in\mathbb{R}^{d\times N}$} is the mapping provided by the gradient w.r.t. the feature vectors.
Given that {\small $\mK_f$} and {\small $\mK_g$} are \acrfull{psd} matrices, we use the Hadamard product {\small $\circ$} between two kernel matrices
\begin{align}
  \mK=\mK_f\circ\mK_g \label{kxk_cov},
\end{align}
to combine them, while preserving positive semi-definiteness.
As {\small $k(\vx, \vx^\prime)=k_f(\vx, \vx^\prime) \cdot k_g(\vx, \vx^\prime)$} is a valid kernel function, {\small $k(\cdot, \cdot)$} also defines a mapping {\small $\vx\rightarrow\psi(\vx)$} which gives {\small $\mK=\Psi^\top\Psi$}. The constructed kernel can be decomposed in the following way:
\begin{align}
  k(\vx_i, \vx_j) &= k_f(\vx_i, \vx_j) \cdot k_g(\vx_i, \vx_j) \nonumber \\
  &= \langle\psi(\vx_i), \psi(\vx_j)\rangle_F \\
  \text{where } \psi(\vx) &= \mathbb{E}_{q(y|\vx)} \nabla_{\vf} \ell(\phi(\vx), y)
  \vf^\top \in\mathbb{R}^{d\times d} . \label{psi-mapping}
\end{align}
Then the implicit mapping is defined as {\small $\psi(\vx)$}.
Specifically, for the input layer, it becomes {\small $\mathbb{E}_{q(y|\vx)}\nabla_\vx \ell(\phi(\vx), y) \vx^\top$}.
The kernel matrix {\small $\mK$} is considered as the \textbf{representation} of the neural network {\small $\phi$} on the dataset $D$.

\subsection{Similarity Index}
Given that the representation is a kernel matrix, it is natural to consider similarity indices that imply the alignment between two kernels. We will consider two potential candidates for alignment metrics --- \acrfull{cka} \citep{Cortes2012AlgorithmsFL} and \acrfull{nbs} \citep{Muzellec2018GeneralizingPE}. Both of them are invariant to any isotropic scaling and rotation in the embedding space when the linear kernel is used, which are desired invariance properties when comparing two models.

\textbf{Centred Kernel Alignment} was empirically illustrated as a more appropriate similarity index when models trained on the same dataset are compared than \acrshort{cca} or linear regression \cite{Kornblith2019SimilarityON}. Suppose two kernel matrices {\small $\mK_1$} and {\small $\mK_2$} are constructed on the same dataset,
and data samples are ordered in the exact same way,
the \acrshort{cka} is defined as:
\begin{align}
  \rho_{CKA}(\mK_1, \mK_2) = \frac{\langle\mK_1,\mK_2\rangle_F}{||\mK_1||_F||\mK_2||_F} \in [0, 1] .
\end{align}
\textbf{Normalised Bures Similarity} normalises the Bures metric described in \citep{Bures1969AnEO} for computing the Wasserstein distance between two elliptical distributions. It is similar to \acrshort{cka} but it was derived from the perspective of optimal transport between distributions. The \acrshort{nbs} is defined as:
\begin{align}
  \rho_{NBS}(\mK_1, \mK_2) = \frac{\Tr(\mK_1^\frac{1}{2}\mK_2\mK_1^\frac{1}{2})^\frac{1}{2}}
  {\sqrt{\Tr(\mK_1)\Tr(\mK_2)}} \in [0, 1].
\end{align}
Both formulae can be simplied as the representation we described above is constructed as a linear kernel $\mK=\Psi^\top \Psi$,
\begin{align}
  \rho_{CKA}(\mK_1, \mK_2) &= \frac{\sum_{i=1}^N\left|\sigma_i^2(\Psi_1\Psi_2^\top)\right|}{||\Psi_1^\top\Psi_1||_F||\Psi_2^\top\Psi_2||_F} \\
  \rho_{NBS}(\mK_1, \mK_2) &= \frac{\sum_{i=1}^N\left|\sigma_i(\Psi_1\Psi_2^\top)\right|}{||\Psi_1||_F||\Psi_2||_F}
\end{align}
where $\sigma_i(\cdot)$ is the $i$-th singular value of a matrix sorted in a descending order according to their magnitudes. Their relationship can be inferred as shown below (refer to the Appendix for a proof):
\begin{align}
  \rho_{CKA}(\mK_1, \mK_2)\leq\rho_{NBS}(\mK_1^2, \mK_2^2)
   \label{ckanbs}
\end{align}

\subsection{Sketching the Kernel Matrix}
Unfortunately, storing the full kernel matrix of individual representations would require {\small $\mathcal{O}(N^2)$} memory whilst matrix multiplication to compute the similarity indices would require {\small $\mathcal{O}(N^3)$} computational time, where {\small $N$} is the number of samples. We wish to ease this computational burden by using a low-rank approximation to the kernel matrices.

There are several plausible approaches to reduce the complexity of computing the similarity, including Nystr{\"o}m method to obtain a rank-{\small $M$} approximation of the representation {\small $\mK$}, where {\small $M \ll N$}.
Here, we outline how sketching can be used to compute a summary {\small $\tilde{\mK}\in\mathbb{R}^{M\times M}$} of the representation {\small $\mK\in\mathbb{R}^{N\times N}$}.
An immediate benefit of using our sketched approach is that
then the complexity of evaluating similarity indicies is reduced from
{\small $\mathcal{O}(N^3)$} to {\small $\mathcal{O}(M^3)$}.
Given a large matrix {\small $\mX \in \mathbb{R}^{N \times d}$}, the property we require
of our sketch is the so-called \emph{subspace embedding} which guarantees that
the column space of {\small $\mX$} is approximately preserved.
For low rank matrices {\small $\mX$}, the column space typically has dimensionality much
smaller than both {\small $N$} and {\small $d$} and consequently we may project down to a
dimensionality proportional to {\small $\mathcal{O}(\text{rank}(\mX)) \ll N$} \citep{udell2019big}.
A comprehensive study of sketching techniques and their theoretical guarantees
can be
found in \citet{Woodruff2014SketchingAA}.
More details on the sketch are provided in Appendix \ref{sec: sketch-details}.

The key concept is to apply a random projection {\small $\mS$} to distribute {\small $N$} data samples into {\small $M$} buckets, then the resulting {\small $M$} buckets as a whole serve as the summary of the original dataset.
There are many options for choosing {\small $\mS$} to guarantee the subspace
embedding property with {\small $\mS$} both dense and sparse, for example, a
suitably scaled matrix with Gaussian random variables.
However, such a dense transform requires generating {\small $Nd$} random variables and
invoking matrix product at a cost of {\small $\mathcal{O}(Nd^2)$} so is not scalable.
For the sake of time and space complexity of our method, we choose to apply
a version of the CountSketch (also known as CWT) of
\citep{Clarkson2013LowRA} which is extremely sparse and can be applied
efficiently by \emph{streaming} through the input.
Specifically, CWT is utilised to speed up
our method, and a detailed description of its application is described in Alg. \ref{alg:cwt}.
The CWT can be seen as applying a sign function to the rows of input {\small $\mX$}
followed by hashing the rows to {\small $M$} buckets uniformly at random.
A detailed definition is given in Appendix \ref{sec: sketch-details}.

\begin{algorithm}
  \caption{Representation Construction with Sketching}
    \algorithmicinput\quad{a neural network {\small $\phi_\theta$},  a dataset {\small $D=\{\vx_i, y\}_{i=1}^N$}, number of buckets for sketching {\small $m$}}, and batch size {\small $bs$} \\
    \algorithmicoutput\quad{two kernel matrices {\small $\tilde{\mK_f}, \tilde{\mK_g}\in\mathbb{R}^{m\times m}$}} \\
    \algorithmicset\quad{{\small $\tilde{\mF}=\mathbf{0}\in\mathbb{R}^{d\times m}$, $\tilde{\mG}=\mathbf{0}\in\mathbb{R}^{d\times m}$, $t=0$}}

    \algorithmicwhile\quad{\small $t \leq \ceil{\frac{N}{bs}}$} \\
    \algorithmicdo \\
    $\text{\hspace{0.5cm}}$ {\small $\mX = \{\vx_i\}_{i=bs\times t+1}^{bs\times (t+1)}$} \\
    $\text{\hspace{0.5cm}}$ feature computation {\small $ \mF = \phi_f(\mX) \in \mathbb{R}^{d\times bs}$} \\
    $\text{\hspace{0.5cm}}$ gradient computation {\small $ \mG = \phi_g(\mX) \in \mathbb{R}^{d\times bs}$} \\
    $\text{\hspace{0.5cm}}$ \algorithmicfor{\{{\small $j=1$}; {\small $j\leq bs$}; {\small $j++$}\}} \\
    $\text{\hspace{1.0cm}}$  sample a binary value {\small $s$} \\
    $\text{\hspace{1.0cm}}$  uniformly sample a bucket and note its index as {\small $k$}  \\
    $\text{\hspace{1.0cm}}$  {\small $\tilde{\mF}[:,k] = \tilde{\mF}[:,k] + s \mF[:,j]$} \\
    $\text{\hspace{1.0cm}}$  {\small $\tilde{\mG}[:,k] = \tilde{\mG}[:,k] + s \mG[:,j]$} \\
    $\text{\hspace{0.5cm}}$\algorithmicendfor \\
    $\text{\hspace{0.5cm}}$ {\small $t = t+1$} \\
    \algorithmicendwhile \\
    {\small $\tilde{\mK_f}=\tilde{\mF}^\top\tilde{\mF}$}, {\small $\tilde{\mK_g}=\tilde{\mG}^\top\tilde{\mG}$}
\label{alg:cwt}
\end{algorithm}

After sketching the two matrices being compared in \acrshort{cka} or
\acrshort{nbs} are individual summaries
{\small $\tilde{\mK_1}=\mS_1^\top\Psi_1^\top\Psi_1\mS_1$} and
{\small $\tilde{\mK_2}=\mS_2^\top\Psi_2^\top\Psi_2\mS_2$} but it is important to
 know how sketching affects the accuracy of similarities.
Here, we provide a bound on the difference between \acrfull{scka} and \acrshort{cka} when the two kernels are evaluated on the same dataset in two steps
(taking $\mS_1 = \mS_2$).
The first step shows how much the numerator changes with sketching, and the second step shows how much the \acrshort{scka} differs from \acrshort{cka}.

\textbf{Lemma } w.p. $1-\delta$
\begin{align}
\left|||\Psi_1^\top\mS^\top\mS\Psi_2||_F - ||\Psi_1^\top\Psi_2||_F \right| \leq \epsilon ||\Psi_1^\top\Psi_2||_F \label{step1}
\end{align}
\textbf{Lemma } w.p. $1-\delta$
\begin{align}
  \left| \rho_\text{SCKA} -  \rho_\text{CKA} \right | &\leq \frac{4\epsilon}{(1-\epsilon)^2} \rho_\text{CKA} \label{step2} \\
  \text{where } \rho_\text{SCKA} & = \rho_\text{CKA}(\tilde{\mK_1}, \tilde{\mK_2}) \nonumber \\
  & = \rho_\text{CKA}(\mS^\top\mK_1\mS, \mS^\top\mK_2\mS),  \nonumber \\
  \text{ and } \rho_\text{CKA} & = \rho_\text{CKA}(\mK_1, \mK_2) \nonumber
\end{align}
The previous two lemmas show that with high probability, sketching preserves the representations learned by the induced kernel.
Beyond merely being fast, sketching also has another critical advantage: we can use the sketched representation which is of fixed size, {\small $M$}, to compare kernels of different shapes. Neither \acrshort{cka} nor \acrshort{nbs} can be applied when the kernel matrices have different shapes -- e.g. the datasets are of different sizes.
The downside is that it is difficult to give a theoretical guarantee on either similarity or on accuracy of the sketched representation.

Sketching is capable of preserving the top principal directions with high probability, and these directions contribute the most in computing the similarity between two kernel matrices in both \acrshort{cka} and \acrshort{nbs}. Therefore, we facilitate similarity indices with sketching to enable comparisons between datasets with different numbers of samples. More generally, our method can be applied to compare different models trained on different datasets.

\subsection{Our Method in a Nutshell}
Given a neural network {\small $\phi$} and a dataset {\small $D$} with {\small $N$} samples, the \textbf{representation} construction step of our method requires only one forward and backward call for each data sample, meanwhile sketches each data sample into {\small $M$} buckets. This step can be parallelised in the batch mode in deep learning frameworks. Then, the inner product of the summary matrix and itself gives the representation {\small $\tilde{\mK}\in\mathbb{R}^{M\times M}$}.
The \textbf{similarity index} takes two representations {\small $\tilde{\mK}_1$} and {\small $\tilde{\mK}_2$} as inputs and outputs a similarity score.

\section{Other Similarity Measures}
Task2Vec and \acrshort{frnorm} are closely related counterparts of ours, and we include them in our experiments. Specifically,
if we denote as {\small $||\theta_1||_\text{fr}$} and {\small $||\theta_2||_\text{fr}$} the individual scalar-representations given by \acrshort{frnorm} for two neural networks, then the distance measure is defined as
{\small $\rho_\text{fr}=\left| ||\theta_1||_\text{fr} - ||\theta_2||_\text{fr} \right|/\sqrt{||\theta_1||_\text{fr}||\theta_2||_\text{fr}}$}.

Since the invariance properties of \acrshort{cka} and \acrshort{nbs} are important in comparing models, we keep similarity indices the same and consider other ways of merging information from feature vectors and gradient vectors. Potential simple combinations include the following methods:
\begin{enumerate}
  \item {\small $\mK=\mK_f + \mK_g$}
  \item {\small $\mK=\Phi_{fg}^\top\Phi_{fg}$}, where {\small $\Phi_{fg}=\Phi_f \circ \Phi_g$}
  \item {\small $\mK=(\frac{1}{2}\mI_N+\frac{1}{2}\mT^{fg}) \mK_f (\frac{1}{2}\mI_N+\frac{1}{2}\mT^{fg})$}, \\
  where {\small $\mT^{fg}=\mK_f^{-\frac{1}{2}} (\mK_f^{\frac{1}{2}} \mK_g \mK_f^{\frac{1}{2}})^\frac{1}{2}\mK_f^{-\frac{1}{2}}$}
\end{enumerate}
Since the resulting matrices of all three aforementioned combinations are \acrshort{psd}, they can be regarded as kernel matrices as well, and the similarity indices described in the previous section can still be applied to compute similarity scores.

\section{Experimental settings and Analysis}

Our experiments illustrate that our method is efficient and informative in indicating the similarity between two trained neural networks. Three image classification datasets, including CIFAR-10, CIFAR-100 \cite{Krizhevsky2009LearningML} and Street View House Number (SVHN) \cite{Netzer2011ReadingDI}, are chosen for comparison. Residual Networks (ResNet) \cite{He2015DeepRL} are used as the architecture for training and comparison after learning. Since a given ResNet is usually a stack of residual blocks and each with two or three convolution layers, to avoid cluttering in our plots, our proposed method and its counterparts are computed at the output of each residual block instead of each convolution layer.

Ten Residual networks with $18$ layers and one with $50$ layers, each with a different random initialisation, are trained on each of the three chosen datasets until convergence, therefore, in total $33$ models are used in our experiments to ensure that observational findings are significant. Training details are described in the supplementary material.

\subsection{Similarity of Independently Trained Models}
\begin{figure*}[t]
    \centering
    \includegraphics[width=0.9\textwidth]{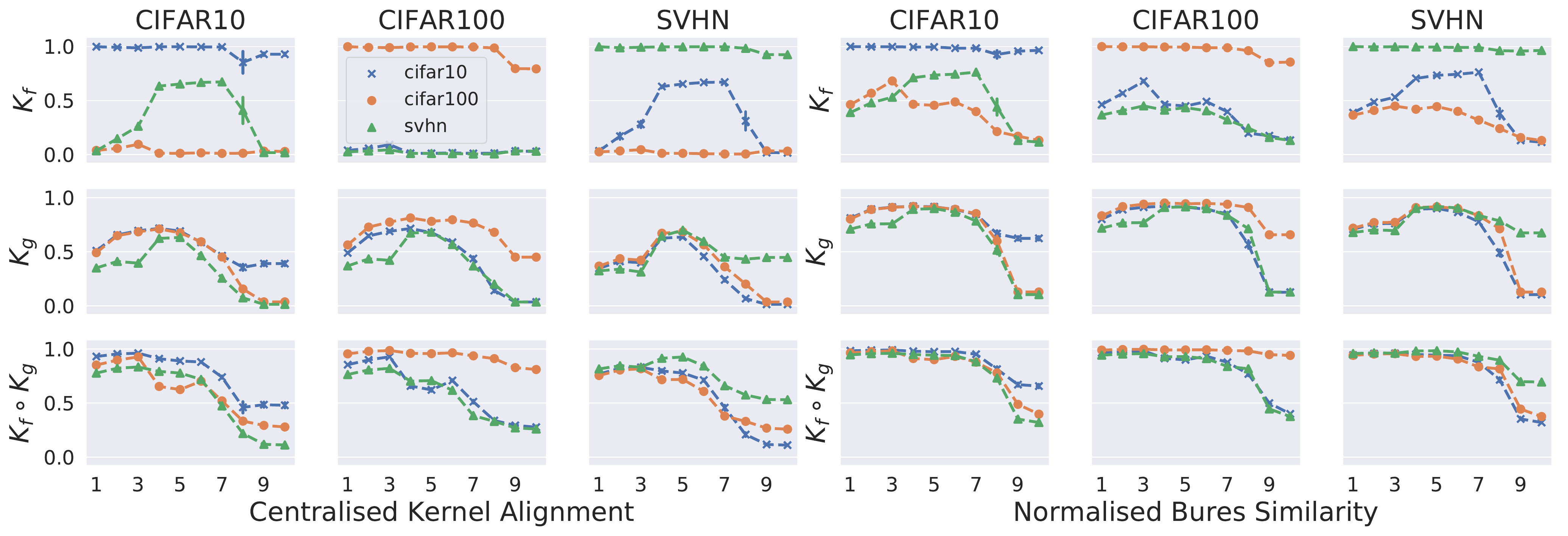}
    \caption{\textbf{Comparison among models trained on individual tasks.} Rows indicate feature-only, gradient-only and our representations respectively. First three columns are using \acrshort{cka} as the similarity index and the last three \acrshort{nbs}. As ten models are trained on each task, each marker is the mean of similarity scores, and each vertical bar represents the standard deviation.
    Combining gradients with the features as the representations helps both similarity indices to give intuitive similarity scores that meet the desiderata described in the main paper. (The x-axis represents the indices of residual blocks, and larger the index, closer the ResBlock is to the output layer.)}
    \label{modelsim}
\end{figure*}
We first consider computing a similarity score at the same ResBlock for two given ResNet-18 models trained on two datasets, and the ideal behaviours of a good comparison method should include:
\begin{enumerate}
  \item assigning a higher similarity score for a pair of models trained on the same dataset but with different initialisations than models trained on different datasets at the same layer;
  \item giving a higher similarity score for any pair of \{CIFAR-10,CIFAR-100\} as they contain natural images and SVHN contains images of digits;
  \item given the previous aspect, being capable of giving a higher similarity score for the pair \{CIFAR-10, SVHN\} than for the pair \{CIFAR-100, SVHN\} since CIFAR-10 and SVHN are both 10-class classification task while CIFAR-100 is for 100-class classification.
\end{enumerate}
The results of our proposed method are presented in Fig. \ref{modelsim} along with feature and gradient-only comparison to illustrate the effectiveness of incorporating gradients. As known, \acrshort{cka} and \acrshort{nbs} are designed for measuring the alignment between two covariance/kernel matrices evaluated on the same dataset, and it gives nonsensical scores when two different datasets are provided. The first row in Fig. \ref{modelsim} which only considers feature vectors into comparison exactly matches our expectation, and it gives high similarity scores for the same dataset, and low and uninterpretable results for different datasets. By incorporating gradient vectors into comparison in our proposed fashion in Eq. \ref{kxk_cov}, which is illustrated in the third row of the same figure, our method is capable of matching all three ideal behaviours described above. The relative comparison results are also consistent between two related similarity indices including \acrshort{cka} and \acrshort{nbs}. Other representations with same similarity indices are reported in Fig. \ref{supp_cka} and \ref{supp_bures} in the supplementary materials.

\begin{figure}[!ht]
  \begin{center}
    \includegraphics[width=0.19\textwidth]{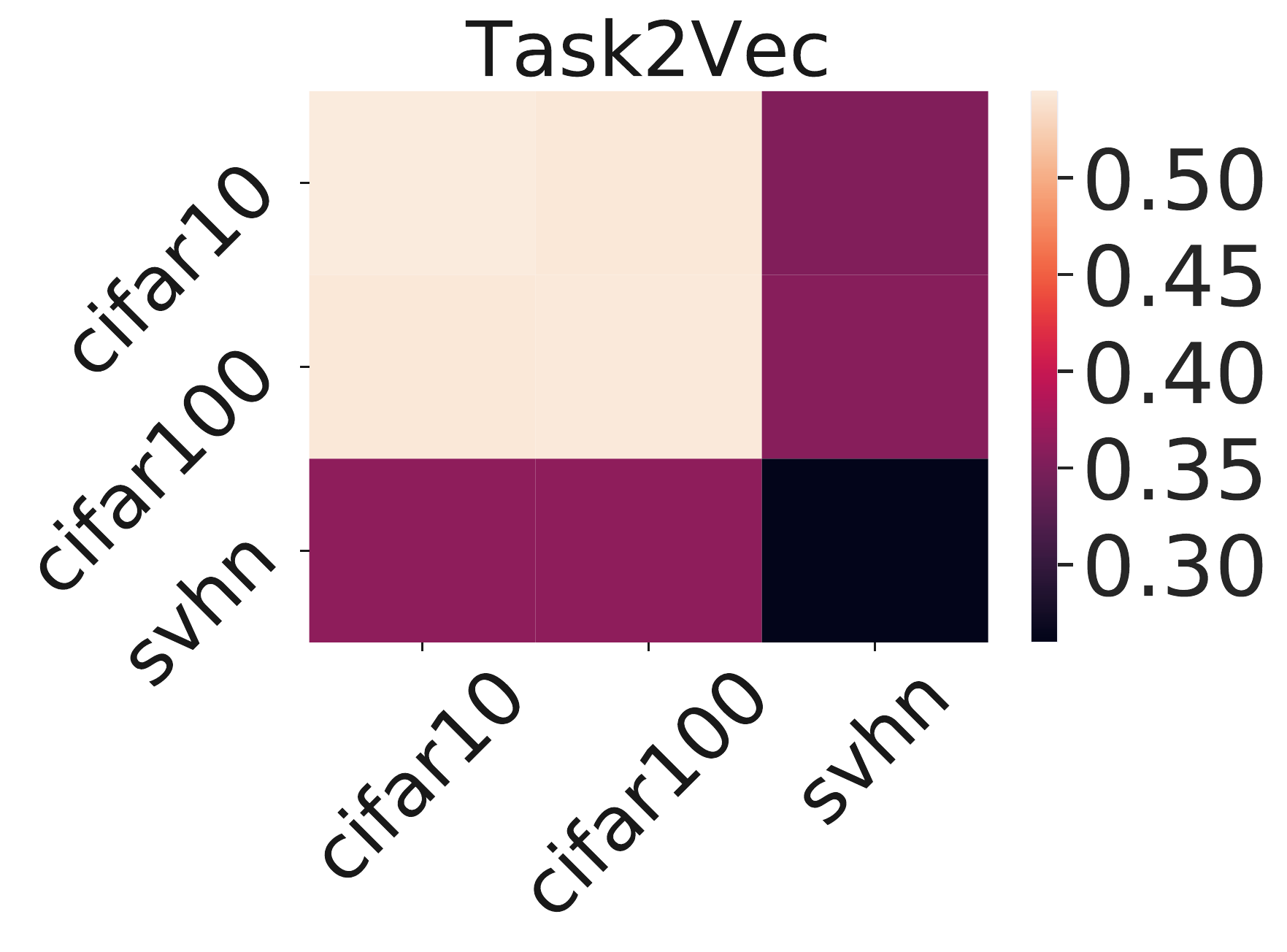}
    \caption{Model comparison using \textbf{Task2Vec}. It gives low similarity scores when comparing between pairs of SVHN models, and scores are even lower than the case when an SVHN model is compared with a CIFAR model.}
    \label{task2vec}
  \end{center}
\end{figure}
\begin{figure}[!ht]
  \begin{center}
    \includegraphics[width=0.43\textwidth]{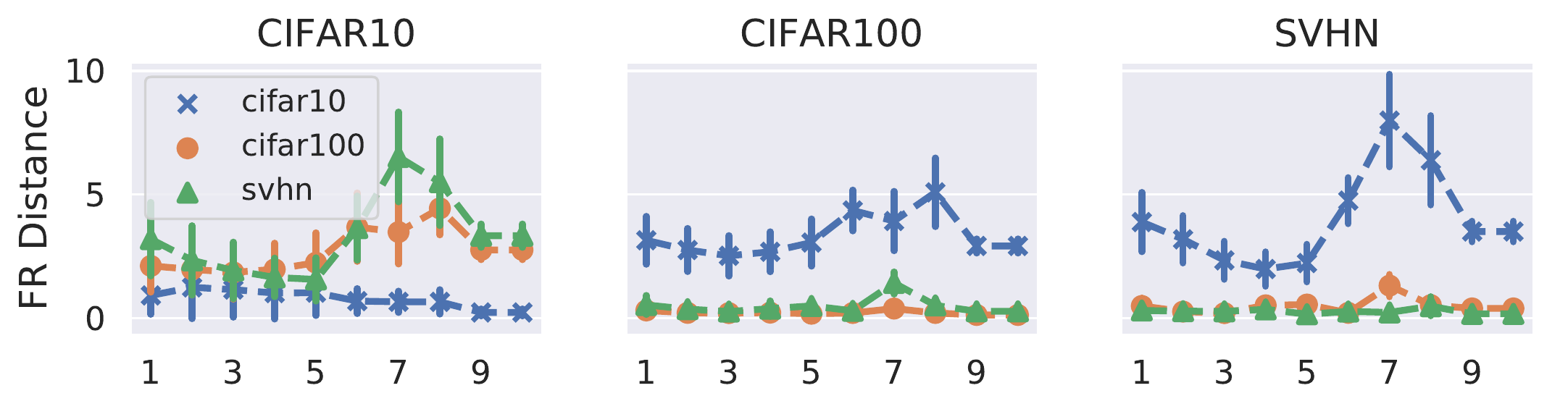}
  \caption{Model comparison using \textbf{Fisher-Rao Distance}. It tends to separate out CIFAR10 models from others, and not able to distinguish CIFAR100 models from SVHN ones.}
  \label{frdis}
  \end{center}
\end{figure}
Task2Vec results are presented in Fig. \ref{task2vec}, and we only computed similarity scores for the entire model instead of individual blocks. As expected, since Task2Vec requires a reference model to compute the \acrshort{fim} for each task, when it is used to compare independently trained models, it doesn't provide insightful similarity scores.
A distance measure based on the \acrshort{frnorm} is presented in Fig. \ref{frdis}, and it does not meet the desiderata. The FR norm was proposed to measure the model complexity of a neural network with minimal assumptions on the loss function and the activation functions, therefore, it encodes little information about the dataset the given neural network was trained on.
\begin{figure*}[t]
  \centering
    \includegraphics[width=0.9\textwidth]{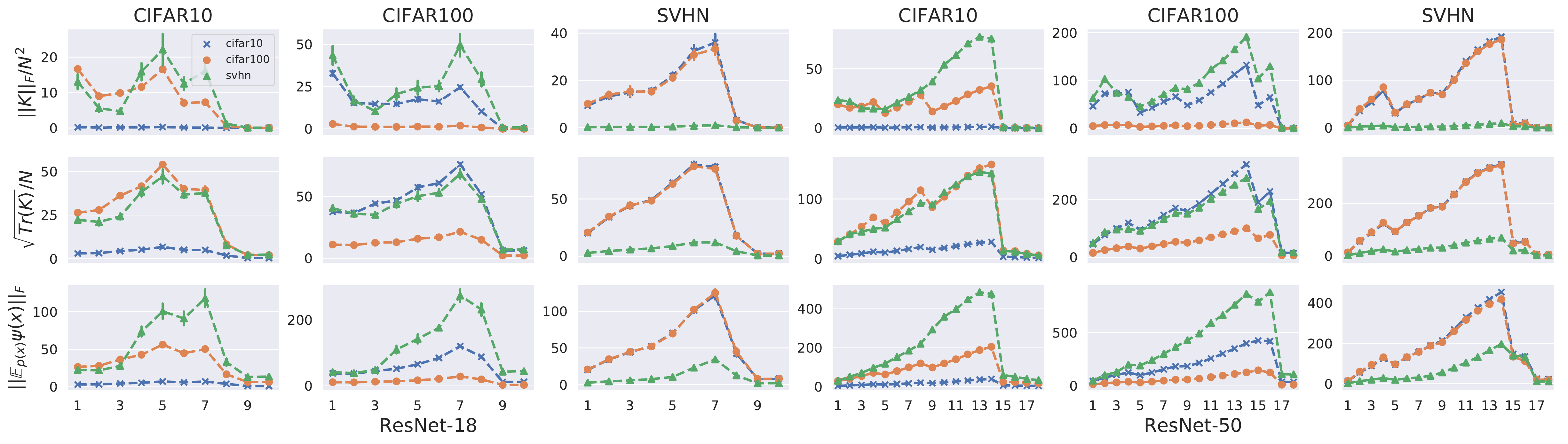}
    \caption{Three quantities {\{\small $||\mK||_F/N^2$ (normalisation factor in CKA),$\sqrt{\Tr(\mK)}/N$ (the factor in NBS),$||\mathbb{E}_{p(\vx)}\psi_\theta(\vx)||_F^2$\}}, which are scaled by the number of layers, computed for a target dataset {\small $D_T$} given a model {\small $\phi\in\{\text{ResNet-18}, \text{ResNet-50}\}$} trained on the source dataset {\small $D_S$}, where {\small $D_S$, $D_T$ $\in \{\text{CIFAR-10}, \text{CIFAR-100}, \text{SVHN}\}$}.
    Titles represent {\small $D_S$} and the legend indicates {\small $D_T$}. The norm {\small $||\mathbb{E}_{p(\vx)}\psi_\theta(\vx)||_F^2$} easily separates out SVHN when the base model is trained on natural images, and other two struggles to achieve.}
    \label{norm}
\end{figure*}

\subsection{Similarity between Individual Residual Blocks}
\begin{figure}[!h]
  \begin{center}
    \includegraphics[width=0.4\textwidth]{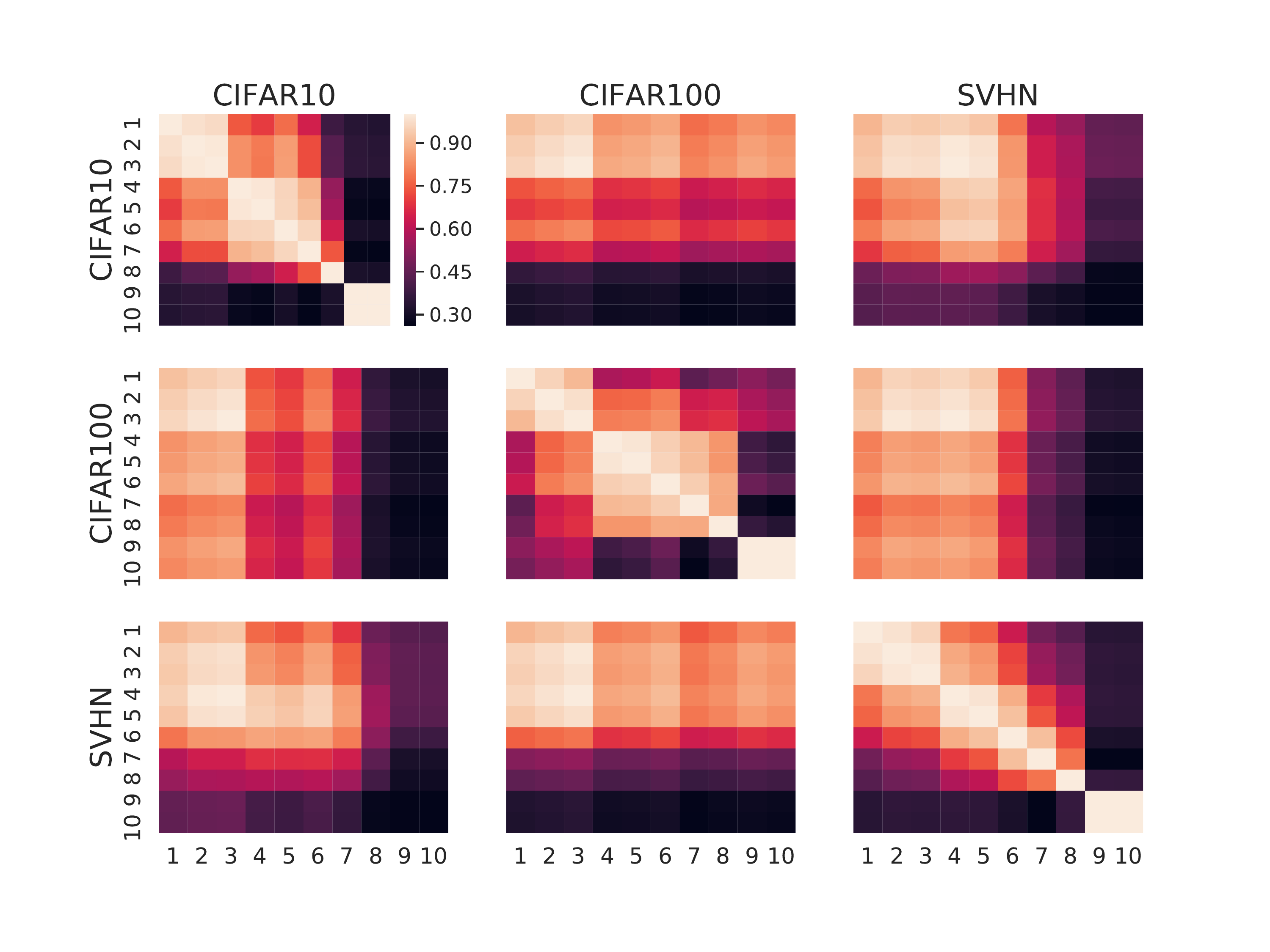}
    \caption{Comparison between every two ResBlocks of ResNet-18 trained on three chosen tasks. The scores are computed using \acrshort{cka} with our representations.}
    \label{layer-wise}
  \end{center}
\end{figure}
A similarity score between any two layers of a trained neural network can inform us the grouping of consecutive layers, which helps distinguish different levels of abstraction in a neural network. In addition, a similarity score between any two layers of two trained models on different datasets gives the rise to question how many layers of parameters learnt on a source task we can transfer to a target task. Fig. \ref{layer-wise} presents a heatmap of similarity scores between every two residual blocks from models trained on two out of three tasks.

It is noticeable that CIFAR-10 and CIFAR-100 have similar behaviours of the similarity across layers, and the critical changes happen at ResBlock3 to ResBlock4 and ResBlock8 to ResBlock9. All layers from CIFAR-100 have high similarity scores to layers before ResBlock8 of models trained on CIFAR-10 and SVHN. This observation implies that functions learnt from harder tasks are more generalisable and transferable to easier tasks. It is also worth noticing that ResBlock9 and 10 of an SVHN model are always not similar to those blocks in CIFAR-10 and CIFAR-100, since the decision boundaries are very different in an SVHN model. Further interpretation might lead to over-interpretation, therefore, we provide more heatmaps in the supplementary materials for readers to discover findings of their interest.

\subsection{Norm of Kernel Mean Embedding}
\label{nkme}
Since the representation is a kernel matrix defined in Eq. \ref{kxk_cov}, and the induced mapping {\small $\psi$} from the kernel is described in Eq. \ref{psi-mapping}, we denote the kernel mean embedding as:
\begin{align}
  \mu_p = \mathbb{E}_{p(\vx)}\psi_\theta(\vx) = \mathbb{E}_{p(\vx)q_\theta(y|\vx)}  \nabla_\vf \ell(\phi(\vx), y) \vf^\top
\end{align}
The embedding {\small $\mu_p$} can be viewed as projecting the data distribution {\small $p(\vx)$} into the \acrshort{rkhs} defined by the kernel {\small $K$} \cite{Muandet2017KernelME}. Since we only have access to the empirical data distribution through the given dataset, the notations here describe the empirical ones.
The norm of the mean embedding is derived by:
\begin{align}
  \hspace{-0.1cm} ||\mu_p|| = ||\mathbb{E}_{p(\vx)q_\theta(y|\vx)}
      \nabla_\vf \ell(\phi(\vx), y) \vf^\top||_F \leq c||\theta||_\text{fr} \label{normkme}
\end{align}
where {\small $||\theta||_\text{fr}$} is the \acrshort{frnorm} of the neural network parametrised by {\small $\theta$}, and {\small $c=1/(L+1)$} and {\small $L$} is the number of layers in the given neural network. The equality holds when {\small $\vf^\top\nabla_\vf \ell(\phi(\vx), y)$} is a constant, and the proof is given in the supplementary material.
\begin{figure*}[t]
  \begin{center}
    \includegraphics[width=0.95\textwidth]{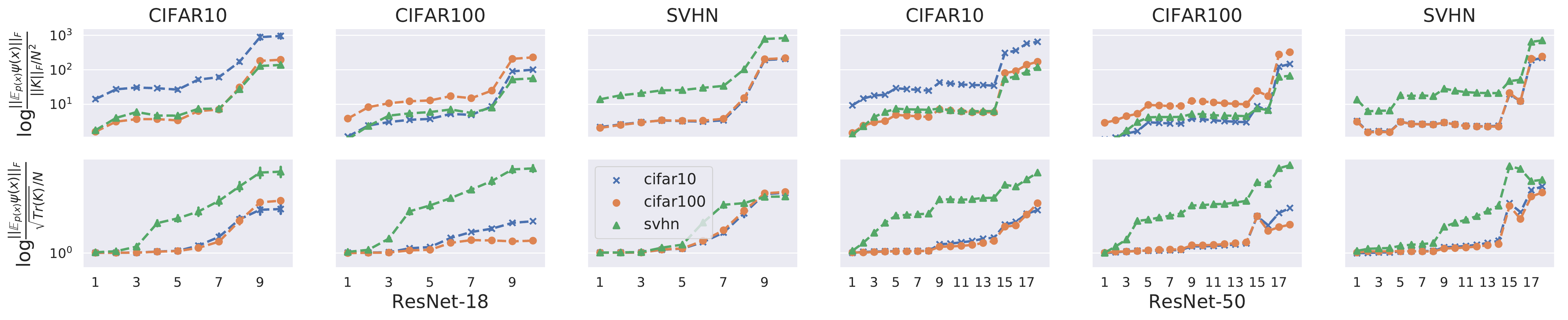}
    \caption{The norm {\small $||\mu_p||$} of our proposed representation divided by the normalisation factors in \acrshort{cka} and \acrshort{nbs} respectively, and plots are in the logarithmic scale. From the perspective of model comparison, the norm with \acrshort{cka} is capable of differentiating $D_S$ from {\small $D_T$}, while it with \acrshort{nbs} always assigns high values to SVHN.}
    \label{div-norm}
  \end{center}
\end{figure*}
The \acrshort{frnorm} \cite{pmlr-v89-liang19a} is described as the low bound of other norms including {\small $\ell_2$}, spectral and path norm of a neural network, which is a direct implication of its property as a measure of model complexity. However, In our case, the scaled {\small $||\mu_p||$} is upper bounded by {\small $||\theta||_\text{fr}$}, and the reduction from the latter quantity to the former one measures the decrease from the model complexity that is independent from the dataset and task to the required complexity to solve the task. In other words, {\small $\mu_p$} depends on both the neural network and the task. Given that we are able to construct the representation - kernel matrix - efficiently through the CWT sketching method, after the kernel matrix is computed, the norm can be easily derived by taking the mean of the kernel matrix and it is
{\small $\mathbb{E}_{\vx_i \sim p(\vx)}\mathbb{E}_{\vx_j\sim p(\vx)} k(\vx_i, \vx_j)=||\mu_p||^2 \label{normcomp}$}.

In order to empirically understand the property of {\small $||\mu_p||$}, we conduct an experiment, in which a dataset $D_T$ is used to compute the norm on a neural network $\phi_S$ trained on a different dataset $D_S$. We would expect three behaviours
\begin{enumerate}
   \item when a dataset $D_T$ is chosen to be the same as the one used to train the neural network which is $D_S$, the norm should be reasonably small;
   \item when the considered two datasets aren't the same, the increase of the norm should imply how much the model needs to adjust itself to adapt to the new dataset.
   \item a neural network with more layers should gives larger norms at individual layers than one with fewer layers.
\end{enumerate}
As the norm {\small $||\mu_p||$} conveys interesting properties about both the trained model and the dataset, it shouldn't be normalised in \acrshort{cka} or \acrshort{nbs}. Therefore, it is worth comparing the norm with the denominators in \acrshort{cka} and \acrshort{nbs} which are {\small $||\mK||_F$} and {\small $\sqrt{\Tr(\mK)}$}. Without losing generality, we rescale {\small $\Psi$} by the number of data samples {\small $N$} in {\small $D_T$} to {\small $\Psi/N$}
in {\small $||\mK||_F/N^2$} and {\small $\sqrt{\Tr(\mK)}/N$}. Fig. \ref{norm} presents three quantities, including the rescaled normalisation factors {\small $||\mK||_F/N^2$}, {\small $\sqrt{\Tr(\mK)}/N$} and our
{\small $||\mu_p||$} evaluated on three previous described datasets and on both ResNet-18 and ResNet-50.

A good sign is that all three quantities have relatively different trends across layers, which means that the information of our interest encoded in {\small $||\mu_p||$} isn't removed during comparison. Our norm is capable of giving a higher value to SVHN than to CIFAR-100 when the chosen neural network was trained on CIFAR-10, while {\small $||\mK||_F/N^2$} struggles and {\ small $\sqrt{\Tr(\mK)}/N$} fails. The visualisation matches our practice that, when finetuning a model trained on CIFAR-10 to {\small $D_T$}, it is easier if $D_T=$CIFAR-100 since they both contain natural images, and the model doesn't need to adjust drastically.

To provide an empirical intuition of the reason of the success of our proposed similarity method, it is critical to see how the norm {\small $||\mu_p||$} changes when being normalised by {\small $||\mK||_F$} and {\small $\sqrt{\Tr(\mK)}$} in \acrshort{cka} and \acrshort{nbs}. Fig. \ref{div-norm} presents the following two quantities:
\begin{align}
  \frac{||\mu_p||}{||\mK||_F/N^2} \text{ and  } \frac{||\mu_p||}{\sqrt{\Tr(\mK)}/N}. \label{normalised}
\end{align}
In Fig. \ref{div-norm}, it shows that the former quantity specificed above gives higher values when $D_S=D_T$, while the latter one always assigns higher values to SVHN as $D_T$ no matter if the model was trained on SVHN or not. For model comparisons, the normalisation factor in \acrshort{nbs} seems to have caused ambiguous results while the one in \acrshort{cka} still gives reasonable illustration of similarity. It implies that \acrshort{cka} is a better similarity index than \acrshort{nbs} at least in our experiments.

\subsection{A Comparison Method Given a Reference Model}

Task2Vec provides a way to compute the similarity between datasets with same number of output classes when a reference model is given, however, the computation cost is high as the gradient computation is w.r.t. the parameters and it is not easily parallelisable.

Through our analysis on {\small $||\mu_p||$}, it is possible to use former quantity in Eq. \ref{normalised} as a similarity score to check how similar a reference model is to a new given dataset. Since the representation size is predefined through the sketching algorithm, and the gradient computation can be easily parallelised in the batch mode, our proposed one has much smaller time and space complexity.

\section{Conclusion}

We argued that to compare the similarity of neural networks it is important to come up with a suitable \textit{representation} of the model and a \textit{similarity index} to compute the similarity scores. We proposed a kernel combination method which combines both feature and gradient vectors as the representation, so that it informs the similarity score with the input data distribution and the model construction.
We augmented well-studied \textit{similarity indices} with sketching techniques to enable comparisons between tasks with data samples of different sizes.

Further analysis on the norm of the kernel mean embedding of the predictive distribution generates insights on understanding the success of our proposed representation, and it also gives a preference on \acrshort{cka} over \acrshort{nbs} as a similarity index. We consider our approach as one step forward in the direction of understanding neural networks and we hope simplicity of our method could facilitate more research ideas in interpreting neural networks.

For future work we would like to explore using this method as a recommender system given large collections of pre-trained models and query datasets.

\subsection*{Acknowledgements}
ST was supported by NSF IIS 1817226 partially. WJM was partially supported by an NSF Graduate Research Fellowship under Grant No. DGE-1839302. CD is supported by European Research Council grant ERC-2014-CoG 647557. We thank Hisham Husain for helpful discussions and Jordan Massiah and Pablo Garcia Moreno for helping preparing the codebase.

\bibliography{example_paper}
\bibliographystyle{icml2020}

\appendix
\onecolumn
\section{Properties of Sketches} \label{sec: sketch-details}
\textbf{Definition}
\textit{CountSketch~\cite{Woodruff2014SketchingAA}}:
Initialise {\small $\mS = \mathbf{0} \in \mathbb{R}^{m \times n}$}.
For every  column {\small $i$} of {\small $\mS$} choose a row $h(i)$ uniformly at random.
Set {\small $\mS_{h(i),i}$} to either $+1$ or $-1$ with equal probability.
Then {\small $\mS = \mB\mD$} with {\small $\mB \in \mathbb{R}^{m \times n}$} is a matrix selecting the
hash buckets assigned to a row of the input and {\small $D$} is a diagonal matrix of
Rademacher random variables.

We also need further definitions:

\textbf{Definition}
\textit{Johnson-Lindenstrauss Transform (JLT)}: A matrix {\small $\mS \in \R^{k \times n}$}
is a \acrshort{JLT} with parameters {\small $\epsilon, \delta, f$}, written
{\small $JLT(\eps, \delta, f)$} if, with probability at least {\small $1-\delta$}, for any
{\small $V \subset \R^n$} of size {\small $f$}:
\begin{align}
\forall \vv, \vv' \in V :
| \langle \vv, \vv' \rangle -  \langle \mS \vv, \mS \vv' \rangle|
\le \epsilon \| \vv \| \|\vv'\|
\end{align}
Our bound applies for any sketch which is a {\small $JLT(\eps, \delta, f)$}.
However, in practice, we instead implement the \acrshort{cwt} as it is
extremely fast to apply.
Technically, the \acrshort{cwt} does not ensure a {\small $JLT(\eps, \delta, f)$} in
the worst case.
This is because it preserves the norm of {\small $f = 2^{\Omega(d)}$} points lying in
a $d$-dimensional subspace rather than an arbitrary set of $f$ points.
Further details can be found on pages 16-17 of \citet{Woodruff2014SketchingAA}.
One sparse transform which achieves our bound is the
\emph{Sparse Johnson-Lindentrauss Transform} of \citet{nelson2013osnap} which
can be thought of as $s$ stacked \acrshort{cwt} matrices of height {\small $m/s$}
for a projection dimension $m$.
Our implementation explored several settings for $s$, however, {\small $s=1$} performed
well-enough and was extremely fast to apply.
This is the reason we use the \acrshort{cwt} in practice but prove our bounds
for general JLT matrices.

In spite of this, there remain some strong theoretical reasons why we expect the
\acrshort{cwt} to perform favourably.
This is because it still preserves all directions which make up the column space
of the input matrix, as formalised below.

\textbf{Definition}
\textit{Subspace Embedding}: A matrix {\small $\mS$} which ensures:
\begin{align}
 (1 - \epsilon)\|\mA\vx\|^2_2 \le \|\mS\mA\vx\|_2^2 \le (1 + \epsilon)\|\mA\vx\|_2^2
\end{align}
is called a \emph{{\small $(1 \pm \epsilon)$} subspace embedding} for the column space
of $A$. In addition, this requirement gives a pointwise guarantee on approximate singular
values.

The main result we need is from \citet{Woodruff2014SketchingAA}:

\textbf{Theorem.} Let {\small $\mA \in \mathbb{R}^{n \times d}$} be a matrix of full rank.
Then for any {\small $\delta \in [0,1]$},  {\small $\mS \in \mathbb{R}^{m \times n}$} is a
{\small $(1 \pm \epsilon)$}-subspace embedding for the column space of {\small $\mA$} with
probability {\small $1-\delta$} provided {\small $m = \mathcal{O}(d^2 / \delta \epsilon^2)$}.
Furthermore, {\small $\mS\mA$} can be computed in time {\small $\mathcal{O}(\text{nnz}(\mA))$}.

Note that the CountSketch has weak failure probability dependence on {\small $\delta$}
but we find in our applications this is not problematic.
Also, depsite the input matrices being dense, it is still quicker to stream
through the input and hash the rows rather than invoking other sketches which
are applied through more expensive methods such as explicit matrix product or
Fast Fourier Transforms.

\section{Training details of Models}
We followed the publically available training procedure of Residual Network, and the details include that (1) the initial learning rate is $0.1$, and it decays by a factor of $10$ every $80$ epochs; (2) each model is trained for $200$ epochs; (3) the optimizer is stochastic gradient descent with batch size $128$ and weight decay coefficient $10^{-4}$ for training each model\footnote{\href{https://github.com/kuangliu/pytorch-cifar}{https://github.com/kuangliu/pytorch-cifar}}.

Model training is done on V100 GPUs, and the main results in our paper including model comparisons and the study on the norm of kernel mean embedding are done on a single Titan 1080 GPU.

\section{Hyperparameters in Sketching}
There are a couple hyperparameters in sketching, including the batch size {\small $bs$} and the sketching size {\small $M$}, which is named as the number of buckets in our main paper. The batch size doesn't have an impact on the results of sketching so we set it to be the largest possible to fit in a single Titan 1080.

The sketching size {\small $M$} is related to the rank of the data matrix. It is computationally difficult to determine or estimate the rank of a given matrix in very high dimension, however, it is well-know that the matrix that contains feature maps generated from neural networks learnt for classification tasks is low-rank. We tested both {\small $M=512$} and {\small $M=128$}, and there is no significant difference between these two settings. In our main paper, all results are presented with {\small $M=512$}.

\section{Proofs}
Here we provide short proofs of inequalities and equalities mentioned in our main paper.

\subsection{Proof of the inequality in Eq. \ref{ckanbs}}
Given the Araki-Lieb-Thirring inequality \citep{araki_inequality_1990} as stated below:
\begin{align}
	&\Tr(\mB\mA\mB)^{rq} \geq \Tr(\mB^r\mA^r\mB^r)^q  , \nonumber \\
	& \text{for } 0\leq r \leq 1, q\geq 0, \text{and any } \mA \succeq 0, \mB \succeq 0
\end{align}
We set {\small $\mA=\mK_2^2$}, {\small $\mB=\mK_1$}, {\small $r=\frac{1}{2}$} and {\small $q=1$} then we have:
\begin{align}
	\Tr(\mK_1\mK_2^2\mK_1)^\frac{1}{2} \geq \Tr(\mK_1^\frac{1}{2}\mK_2\mK_1^\frac{1}{2}) = \langle\mK_1,\mK_2\rangle_F
\end{align}
The first inequality can be derived as follows:
\begin{align}
	\rho_\text{NBS}(\mK_1^2, \mK_2^2)=&\frac{\Tr(\mK_1\mK_2^2\mK_1)^\frac{1}{2}}{\sqrt{\Tr(\mK_1^2)\Tr(\mK_2^2)}} =\frac{\Tr(\mK_1\mK_2^2\mK_1)^\frac{1}{2}}{||\mK_1||_F||\mK_2||_F}
	\geq \frac{\langle\mK_1,\mK_2\rangle_F}{||\mK_1||_F||\mK_2||_F} = \rho_\text{CKA}(\mK_1, \mK_2)
\end{align}

\subsection{Proof of the inequality in Eq. \ref{normkme}}
It is straightforward given the Jensen's inequality:
\begin{align}
	&||\mu_p||^2 = ||\mathbb{E}_{p(\vx)q_\theta(y|\vx)}
  \nabla_\vf \ell(\phi(\vx), y) \vf^\top||_F^2 \nonumber \\
  \leq& \mathbb{E}_{p(\vx)q_\theta(y|\vx)}||\nabla_\vf \ell(\phi(\vx), y) \vf^\top||_F^2  \text{(Jensen's inequality)}\nonumber \\
  =& \mathbb{E}_{p(\vx)q_\theta(y|\vx)}||\vf^\top\nabla_\vf \ell(\phi(\vx), y)||_F^2
  = \frac{||\theta||^2_\text{fr}}{(L+1)^2}
\end{align}
Therefore, {\small $||\mu_p|| \leq ||\theta||_\text{fr}/(L+1)$}.

\subsection{Proof of the equality in Eq. \ref{normcomp}}
The computation of {\small $\mu_p$} relies on the following equality:
\begin{align}
	&\mathbb{E}_{\vx_i \sim p(\vx)}\mathbb{E}_{\vx_j\sim p(\vx)} k(\vx_i, \vx_j) = \langle\mathbb{E}_{\vx_i \sim p(\vx)}\psi_\theta(\vx_i), \mathbb{E}_{\vx_j \sim p(\vx)}\psi_\theta(\vx_j)\rangle_F
  = ||\mathbb{E}_{p(\vx)}\psi_\theta(\vx)||_F^2=||\mu_p||^2
\end{align}
Therefore, we can compute {\small $||\mu_p||$} very efficiently once the kernel matrix is constructed.

\subsection{Proof of the bound in Eq. \ref{step1}}
We slightly alter the notations to match the convention used in \citet{Woodruff2014SketchingAA}. Here, we set {\small $\Psi_1\in\mathbb{R}^{N\times d_1}$}, {\small $\Psi_2\in\mathbb{R}^{N\times d_2}$} and {\small $\mS\in\mathbb{R}^{M\times N}$}.
Let {\small $\mS$} be a {\small $JLT(\eps, \delta, f)$}.

Given in \citet{Woodruff2014SketchingAA} that
\begin{align}
  \Pr\left[\left|\langle \mS(\Psi_1)_{\cdot i}, \mS(\Psi_2)_{\cdot j} \rangle - \langle (\Psi_1)_{\cdot i}, (\Psi_2)_{\cdot j} \rangle \right| \leq \epsilon ||(\Psi_1)_{\cdot i}||_2||(\Psi_2)_{\cdot j}||_2 \right] \geq 1-\delta
\end{align}

It can be written as:
\begin{align}
  \langle \mS(\Psi_1)_{\cdot i}, \mS(\Psi_2)_{\cdot j} \rangle &= \langle (\Psi_1)_{\cdot i}, (\Psi_2)_{\cdot j} \rangle \pm \epsilon ||(\Psi_1)_{\cdot i}||_2||(\Psi_2)_{\cdot j}||_2 \\
  \langle \mS(\Psi_1)_{\cdot i}, \mS(\Psi_2)_{\cdot j} \rangle^2 &= \left(\langle (\Psi_1)_{\cdot i}, (\Psi_2)_{\cdot j} \rangle \pm \epsilon ||(\Psi_1)_{\cdot i}||_2||(\Psi_2)_{\cdot j}||_2\right)^2 \\
  & = \langle (\Psi_1)_{\cdot i}, (\Psi_2)_{\cdot j} \rangle ^2
  + \epsilon^2 ||(\Psi_1)_{\cdot i}||^2_2||(\Psi_2)_{\cdot j}||^2_2
  \pm 2\epsilon \langle (\Psi_1)_{\cdot i}, (\Psi_2)_{\cdot j} \rangle ||(\Psi_1)_{\cdot i}||_2||(\Psi_2)_{\cdot j}||_2
\end{align}

According to Cauchy-Schwartz inequality, we have {\small $\langle (\Psi_1)_{\cdot i}, (\Psi_2)_{\cdot j} \rangle \leq ||(\Psi_1)_{\cdot i}||_2||(\Psi_2)_{\cdot j}||_2$}. \\
Then the upper bound can be simplified as
\begin{align}
  \langle \mS(\Psi_1)_{\cdot i}, \mS(\Psi_2)_{\cdot j} \rangle^2 &\leq \langle (\Psi_1)_{\cdot i}, (\Psi_2)_{\cdot j} \rangle ^2
  + \epsilon^2 ||(\Psi_1)_{\cdot i}||^2_2||(\Psi_2)_{\cdot j}||^2_2
  + 2\epsilon \langle (\Psi_1)_{\cdot i}, (\Psi_2)_{\cdot j} \rangle ||(\Psi_1)_{\cdot i}||_2||(\Psi_2)_{\cdot j}||_2 \\
  &\leq \langle (\Psi_1)_{\cdot i}, (\Psi_2)_{\cdot j} \rangle ^2
  + \epsilon^2 ||(\Psi_1)_{\cdot i}||^2_2||(\Psi_2)_{\cdot j}||^2_2
  + 2\epsilon ||(\Psi_1)_{\cdot i}||^2_2||(\Psi_2)_{\cdot j}||^2_2 \\
  &\leq \langle (\Psi_1)_{\cdot i}, (\Psi_2)_{\cdot j} \rangle ^2
  + (\epsilon^2 + 2\epsilon) ||(\Psi_1)_{\cdot i}||^2_2||(\Psi_2)_{\cdot j}||^2_2
\end{align}

Similarly, the lower bound can be simplified as:
\begin{align}
  \langle \mS(\Psi_1)_{\cdot i}, \mS(\Psi_2)_{\cdot j} \rangle^2 &\geq \langle (\Psi_1)_{\cdot i}, (\Psi_2)_{\cdot j} \rangle ^2
  + \epsilon^2 ||(\Psi_1)_{\cdot i}||^2_2||(\Psi_2)_{\cdot j}||^2_2
  - 2\epsilon \langle (\Psi_1)_{\cdot i}, (\Psi_2)_{\cdot j} \rangle ||(\Psi_1)_{\cdot i}||_2||(\Psi_2)_{\cdot j}||_2 \\
  &\geq \langle (\Psi_1)_{\cdot i}, (\Psi_2)_{\cdot j} \rangle ^2
  + \epsilon^2 ||(\Psi_1)_{\cdot i}||^2_2||(\Psi_2)_{\cdot j}||^2_2
  - 2\epsilon ||(\Psi_1)_{\cdot i}||^2_2||(\Psi_2)_{\cdot j}||^2_2 \\
  &\geq \langle (\Psi_1)_{\cdot i}, (\Psi_2)_{\cdot j} \rangle ^2
  + (\epsilon^2 - 2\epsilon) ||(\Psi_1)_{\cdot i}||^2_2||(\Psi_2)_{\cdot j}||^2_2
\end{align}

Now we have:
\begin{align}
  \langle \mS(\Psi_1)_{\cdot i}, \mS(\Psi_2)_{\cdot j} \rangle^2 = \langle (\Psi_1)_{\cdot i}, (\Psi_2)_{\cdot j} \rangle ^2
  + (\epsilon^2 \pm 2\epsilon) ||(\Psi_1)_{\cdot i}||^2_2||(\Psi_2)_{\cdot j}||^2_2
\end{align}

We sum over {\small $d_1$} columns in {\small $\Psi_1$} and {\small $d_2$} columns in {\small $\Psi_2$}
\begin{align}
  \sum_{i=1}^{d_1}\sum_{j=1}^{d_2} \langle \mS(\Psi_1)_{\cdot i}, \mS(\Psi_2)_{\cdot j} \rangle^2  & = \sum_{i=1}^{d_1}\sum_{j=1}^{d_2}\langle (\Psi_1)_{\cdot i}, (\Psi_2)_{\cdot j} \rangle ^2
  + \sum_{i=1}^{d_1}\sum_{j=1}^{d_2} (\epsilon^2 \pm 2\epsilon) ||(\Psi_1)_{\cdot i}||^2_2||(\Psi_2)_{\cdot j}||^2_2 \\
  ||\Psi_1^\top\mS^\top\mS\Psi_2||_F^2 &= ||\Psi_1^\top\Psi_2||_F^2 + (\epsilon^2 \pm 2\epsilon) ||\Psi_1||_F^2||\Psi_2||_F^2
\end{align}

Since {\small $||\Psi_1||_F^2||\Psi_2||_F^2 \geq ||\Psi_1^\top\Psi_2||_F^2$}, the bound can be tighten by
\begin{align}
  ||\Psi_1^\top\mS^\top\mS\Psi_2||_F^2 &= (1\pm\epsilon)^2||\Psi_1^\top\Psi_2||_F^2
\end{align}

Therefore,
\begin{align}
  \Pr\left[\left|||\Psi_1^\top\mS^\top\mS\Psi_2||_F - ||\Psi_1^\top\Psi_2||_F \right| \leq \epsilon ||\Psi_1^\top\Psi_2||_F \right] \geq 1-\delta
\end{align}

\subsection{Proof of the bound in Eq. \ref{step2}}
By setting {\small $\Psi_1=\Psi_2=\Psi$} we can easily derive
\begin{align}
  \Pr\left[\left|||\Psi^\top\mS^\top\mS\Psi||_F - ||\Psi^\top\Psi||_F \right| \leq \epsilon ||\Psi^\top\Psi||_F \right] \geq 1-\delta
\end{align}

Then we have:
\begin{align}
  ||\Psi_1^\top\mS^\top\mS\Psi_1||_F||\Psi_2^\top\mS^\top\mS\Psi_2||_F &= (1\pm\epsilon)^2||\Psi_1^\top\Psi_1||_F||\Psi_2^\top\Psi_2||_F
\end{align}

Therefore,
\begin{align}
  & \left(\frac{1-\epsilon}{1+\epsilon}\right)^2 \rho_\text{CKA} \leq \rho_\text{SCKA} \leq \left(\frac{1+\epsilon}{1-\epsilon}\right)^2 \rho_\text{CKA} \\
  \text{where } & \rho_\text{SCKA}=\frac{||\Psi_1^\top\mS^\top\mS\Psi_2||_F^2}{||\Psi_1^\top\mS^\top\mS\Psi_1||_F||\Psi_2^\top\mS^\top\mS\Psi_2||_F}
  \text{ and }  \rho_\text{CKA}=\frac{||\Psi_1^\top\Psi_2||_F^2}{||\Psi_1^\top\Psi_1||_F||\Psi_2^\top\Psi_2||_F}
\end{align}

Then the absolute difference between the Sketched CKA and the original CKA is bounded as shown below:
\begin{align}
  \left| \rho_\text{SCKA} - \rho_\text{CKA}\right | \leq \max\{\frac{4\epsilon}{(1+\epsilon)^2},\frac{4\epsilon}{(1-\epsilon)^2}\} \rho_\text{CKA} = \frac{4\epsilon}{(1-\epsilon)^2} \rho_\text{CKA}
\end{align}

Formally,
\begin{align}
  \Pr\left[\left| \rho_\text{SCKA} - \rho_\text{CKA}\right | \leq \frac{4\epsilon}{(1-\epsilon)^2} \rho_\text{CKA} \right] \geq 1-\delta
\end{align}

\clearpage
\section{Additional figures}
\subsection{Model Comparison with different representations}

\begin{figure*}[h]
  \centering
  \includegraphics[width=\textwidth]{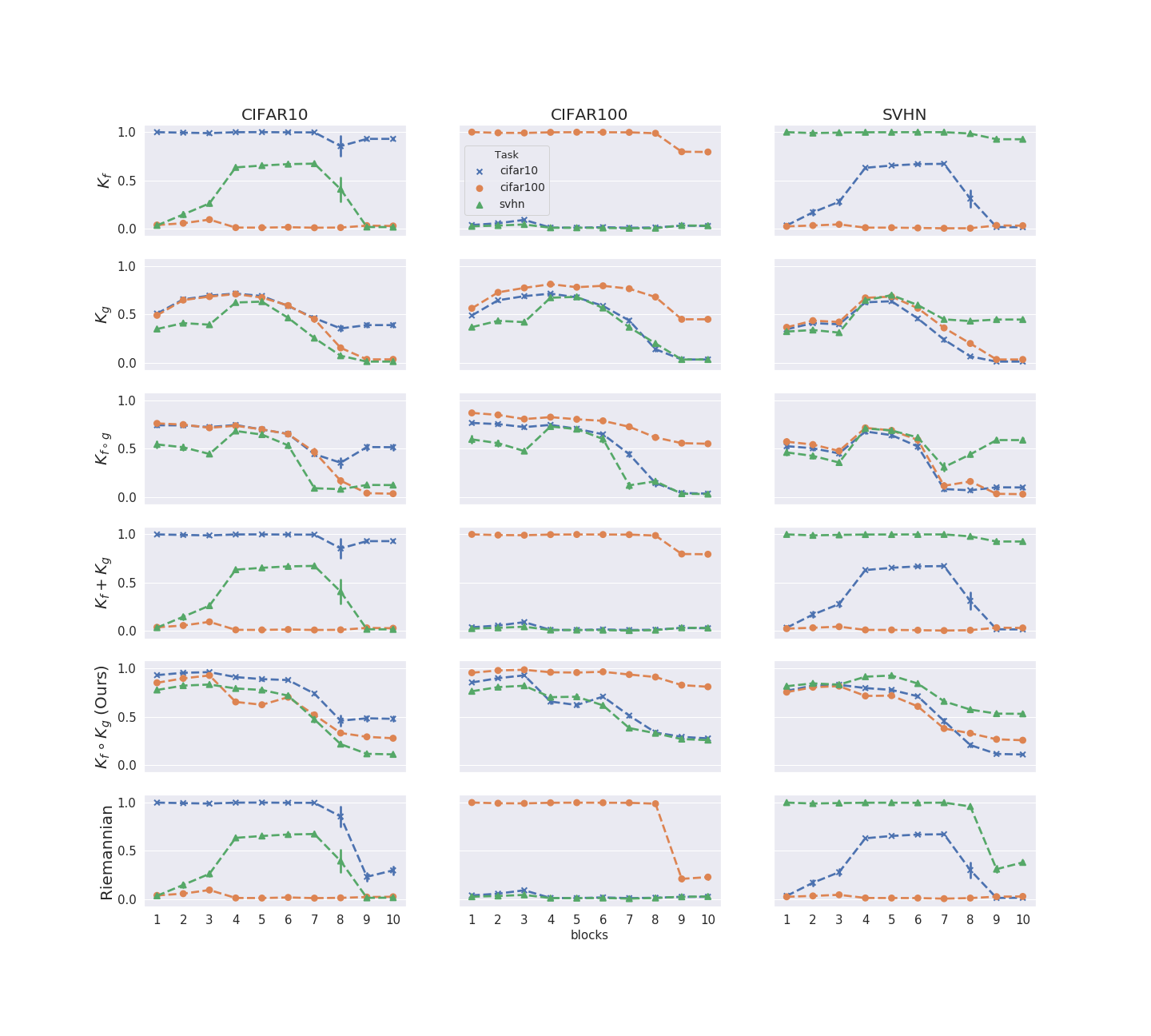}
  \caption{Comparing independently trained models using CKA with different representations. It is clear that our proposed representation gives the illustration that matches the desiderata.}
  \label{supp_cka}
\end{figure*}

\begin{figure*}[t]
  \centering
  \includegraphics[width=\textwidth]{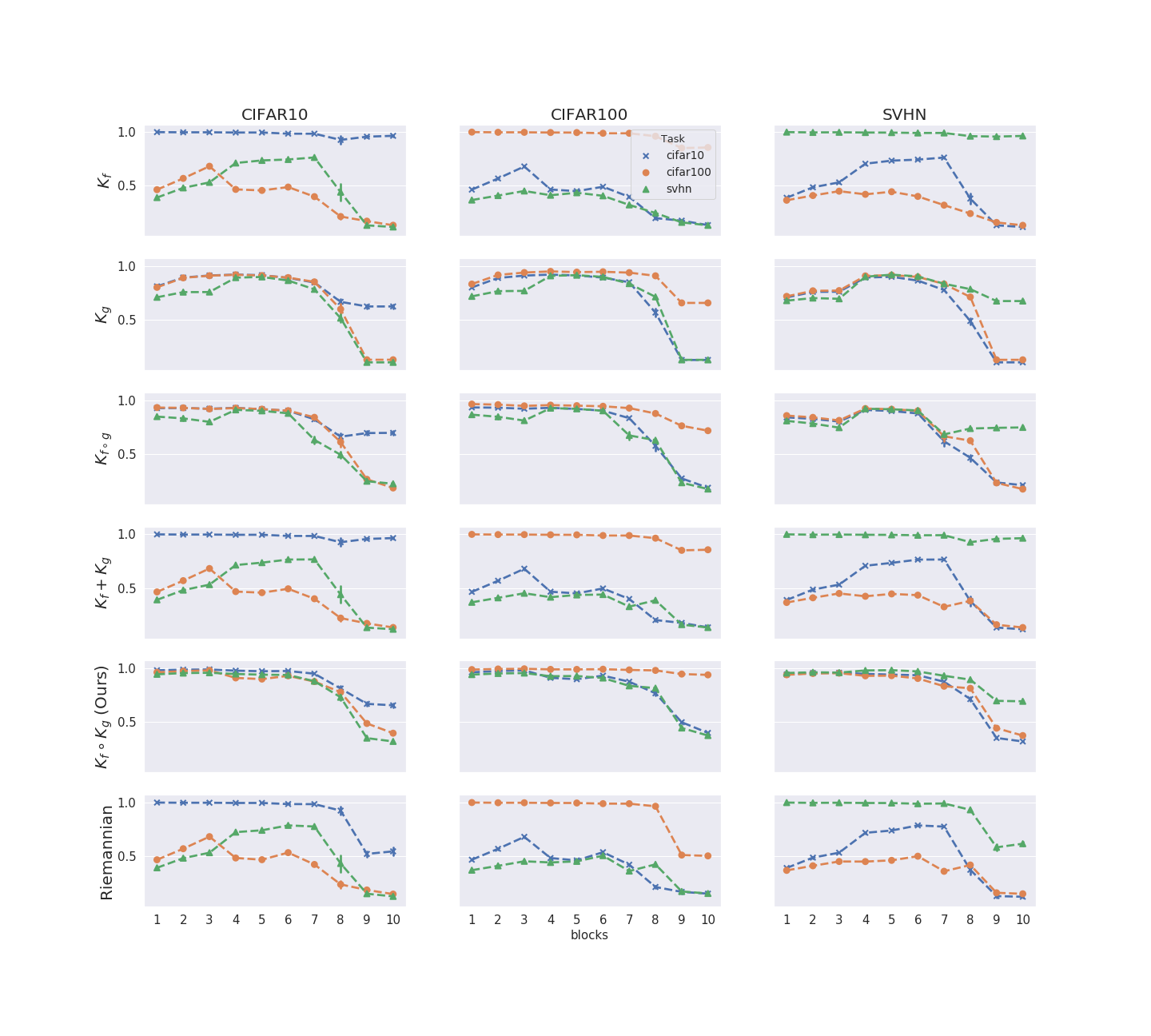}
  \caption{Comparing independently trained models using NBS with different representations.}
  \label{supp_bures}
\end{figure*}

\clearpage
\subsection{Similarity between Individual Residual Blocks}
\subsubsection{Feature-only Representation}
\begin{figure}[h]
  \begin{center}
    \includegraphics[width=\textwidth]{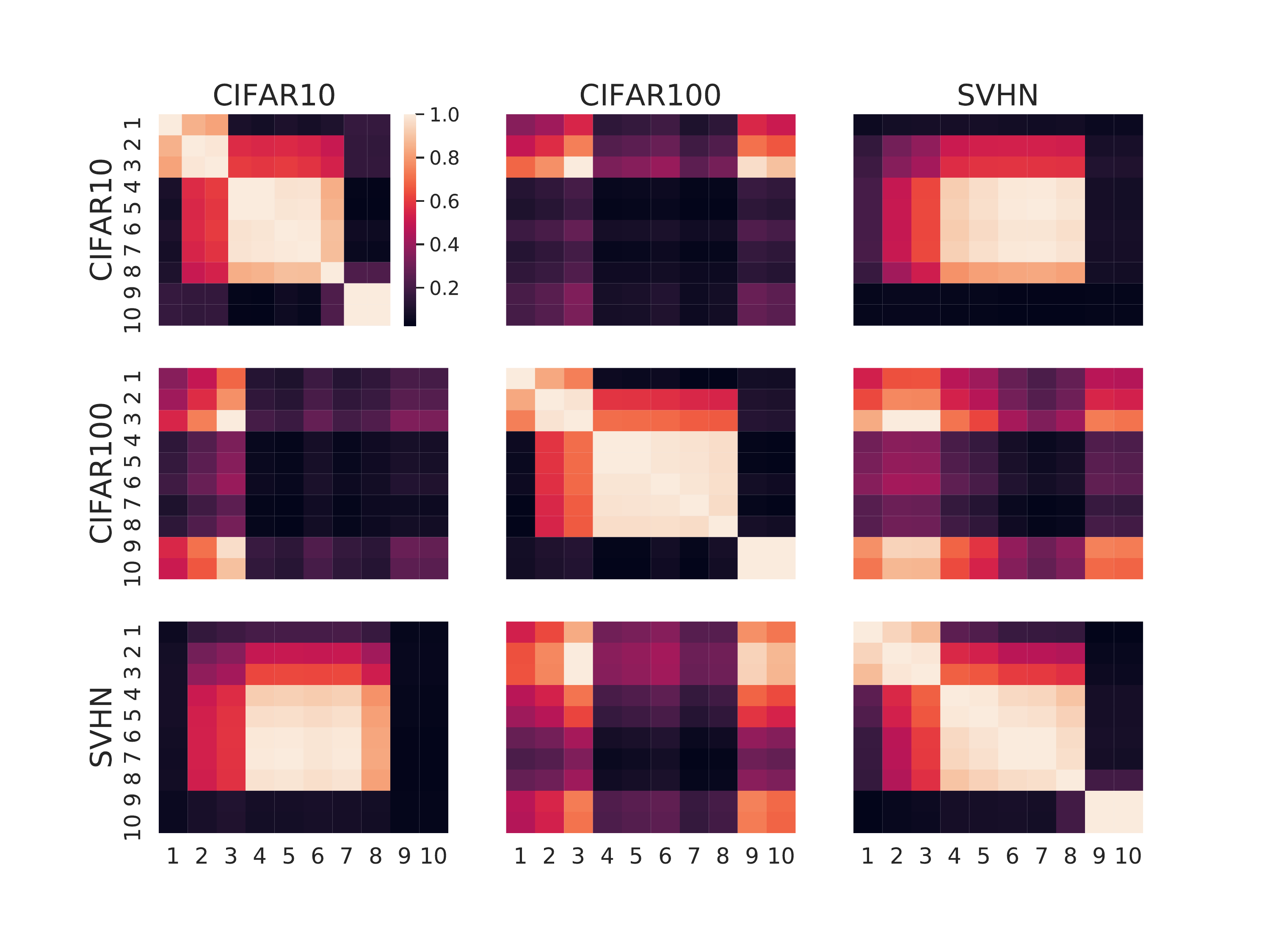}
    \caption{Comparison based on feature vectors only between every two residual blocks of ResNet-18 trained on three chosen tasks. The scores are computed using CKA.}
    \label{18-layers-feat}
  \end{center}
\end{figure}

\clearpage
\subsubsection{Gradient-only Representation}
\begin{figure}[h]
  \begin{center}
    \includegraphics[width=\textwidth]{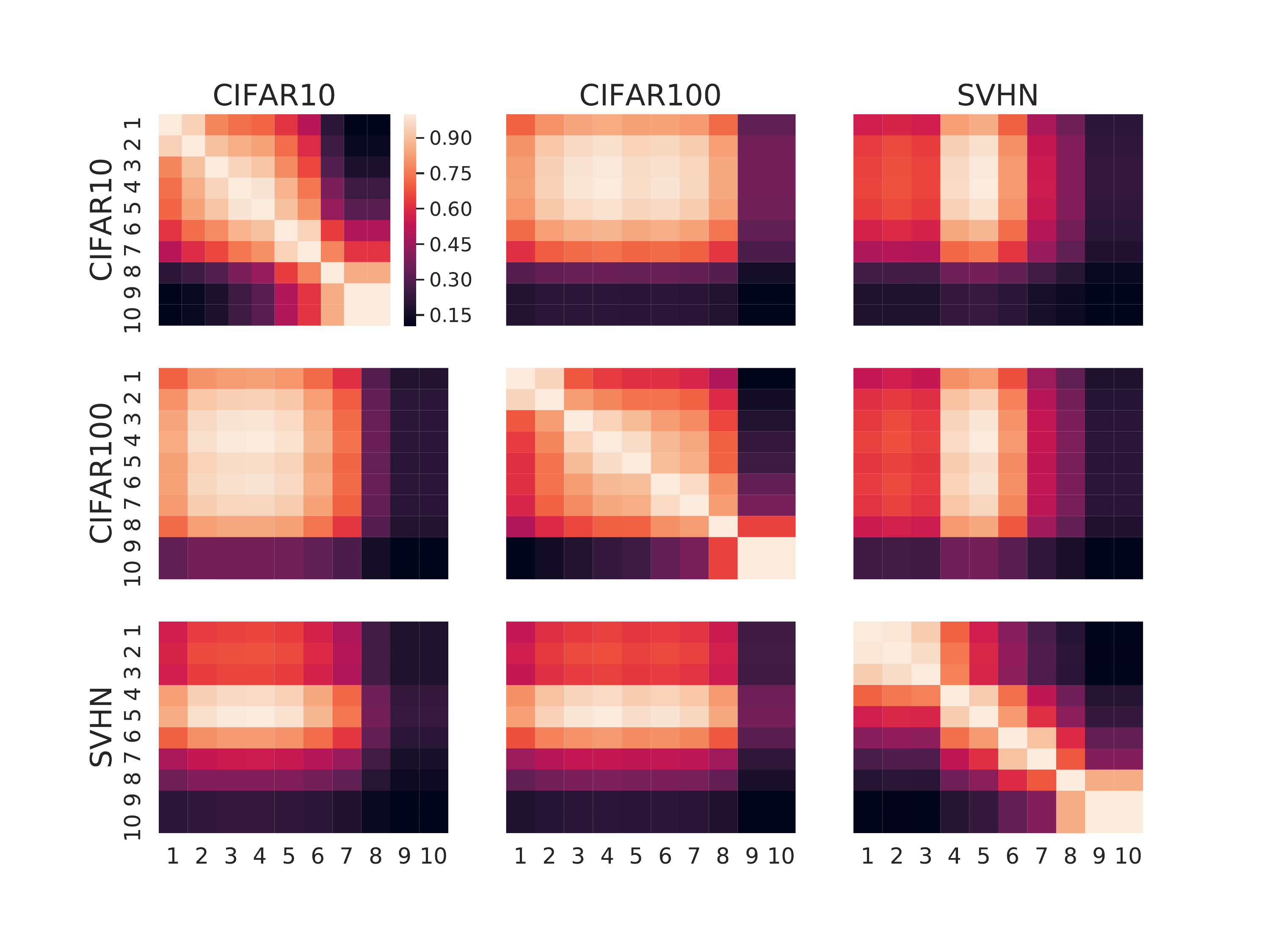}
    \caption{Comparison based on gradient vectors only between every two residual blocks of ResNet-18 trained on three chosen tasks. The scores are computed using CKA.}
    \label{18-layers-grad}
  \end{center}
\end{figure}

\clearpage
\subsubsection{Our Representation with Features and Gradients}
\begin{figure}[h]
  \begin{center}
    \includegraphics[width=\textwidth]{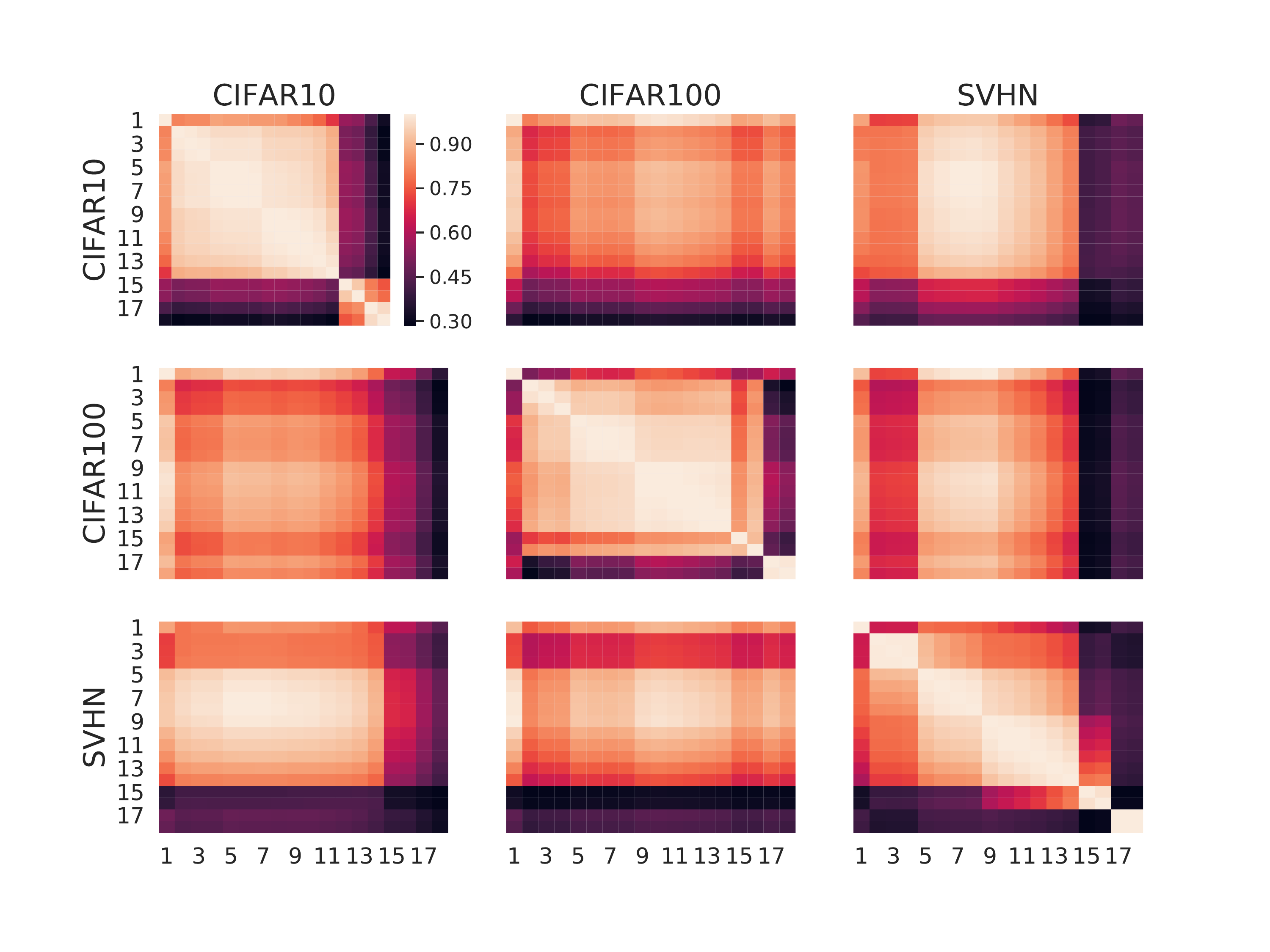}
    \caption{Comparison based on our proposed representations only between every two residual blocks of ResNet-50 trained on three chosen tasks.}
    \label{50-layers-kfokg}
  \end{center}
\end{figure}

\begin{figure}[h]
  \begin{center}
    \includegraphics[width=\textwidth]{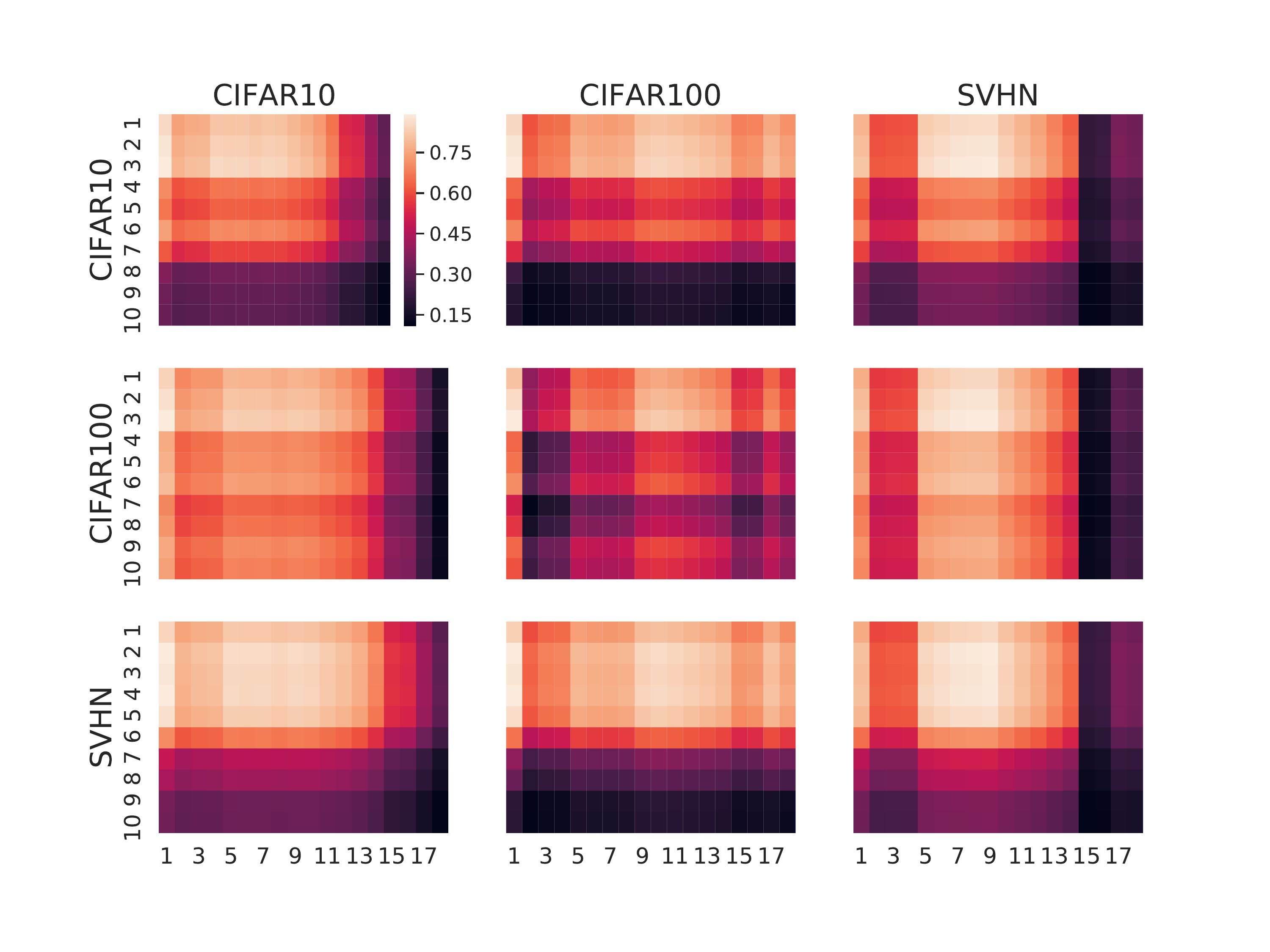}
    \caption{Comparison based on our proposed representations only between every two residual blocks, in which one is from ResNet-18 and the other ResNet-50, trained on three chosen tasks.}
    \label{18-50-layers-kfokg}
  \end{center}
\end{figure}

\clearpage
\section{Pretrained ImageNet models as the base model}
As mentioned in Section \ref{nkme}, the value, which is the norm of the empirical kernel mean embedding evaluated on a dataset given a base model {\small $\mu_p$} divided by the norm of the empirical kernel matrix evaluated on the same dataset {\small $||\mK||_F$}, could serve as a scalar-reference for how much the base model, potentially pretrained, needs to adapt to the given dataset.
\begin{align}
  \log \frac{||\mu_p||}{||\mK||_F/N^2} \label{the-scalar}
\end{align}
Here, the pretrained ImageNet models, specifically ResNet models with different numbers of residual blocks, including ResNet-152, 101, 50, 34 and 18, are selected as the base models, and the same three datasets used in the main paper, which are CIFAR10, CIFAR100 and SVHN, are also used here to demonstrate the former scalar's capability in distinguishing tasks given a base model.

\begin{figure}[t]
  \centering
  \includegraphics[width=\textwidth]{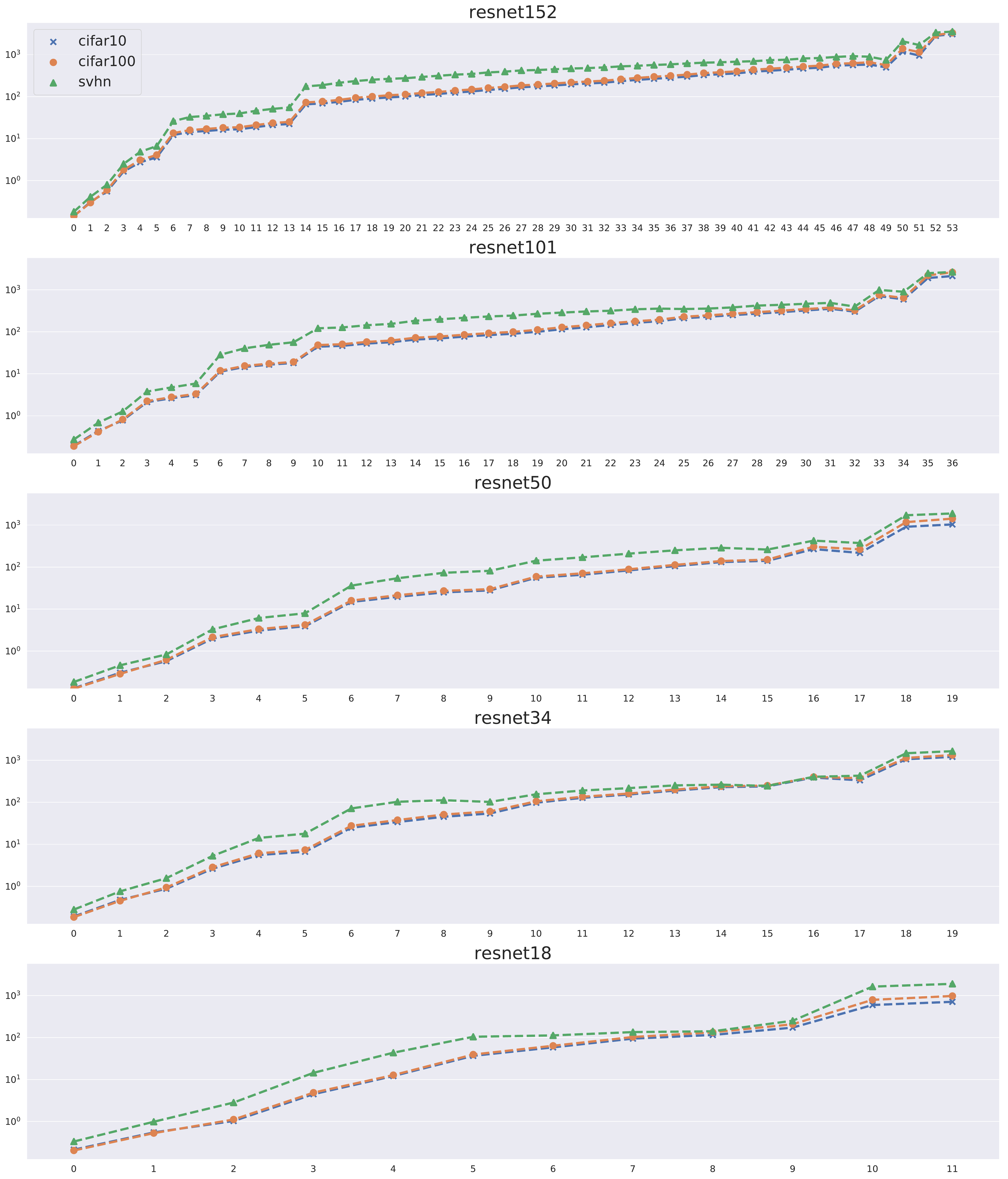}
  \caption{The scalar in \ref{the-scalar} evaluated on three tasks on pretrained ImageNet models. The x-axis indicates the layer index, and larger the number, closer the layer is to the output layer. It is clear presented that, across models and layers, SVHN gives larger values than CIFAR10 and CIFAR100 do, and CIFAR100 gives similar values as CIFAR10 does until higher layers where CIFAR100 gives larger values.}
  \label{scalar-imagenet}
\end{figure}

Fig. \ref{scalar-imagenet} shows the scalars of individual layers in \ref{the-scalar} evaluated on three tasks given a ResNet model pretrained on ImageNet dataset. The scalar can serve as an indicator on how much the model has to adjust to adapt to a new task.

\clearpage
\section{Kernel Ridge Regression as Classification}
One potential application of our proposed method is that the sketched kernel matrix could be used to make predictions on a new task without finetuning the pretrained model. For example, one could apply a ResNet model on pretrained ImageNet dataset on tasks including CIFAR10, STL10, etc., without finetuning the model. Since our method sketches down the number of data samples in a new task to a fixed number to reduce the memory complexity for further analysis, it becomes critical to consider the corresponding transformation of the labels in the training set in terms of making predictions.

Here, we investigate the possibility of solving classification tasks using kernel ridge regression. Given a dataset {\small $D=\{(\vx_i, y_i)\}_{i=1}^N$}, we first convert each label to a onehot vector {\small $\vt_i=[(t_i)_1, (t_i)_2, ..., (t_i)_C]$}, where {\small $C$} is the total number of classes, and {\small $(t_i)_k=1$} when {\small $k=y_i$} and {\small $0$} otherwise. Now, given the dataset
{\small $D=\{(\vx_i, \vt_i)\}_{i=1}^N$} and a pretrained model {\small $\phi$}, our method sketches {\small $\vx_i$} and {\small $\vt_i$} with the same random projection matrix, and provides a sketched kernel matrix {\small $\tilde{\mK}\in\mathbb{R}^{M \times M}$} and a target matrix
{\small $\tilde{\mT}\in\mathbb{R}^{C \times M}$}. Once a new data sample {\small $\vx^\star$} is given, the prediction is made by the following formula
\begin{align}
  \vt^\star &= \tilde{\mT}(\tilde{\mK} + \alpha\mI)^{-1}\vk(\vx^\star), \\
  y^\star &= \argmax_{k}\vt^\star_k
\end{align}
where {\small $\alpha$} is the strength of the functional regularisation whose optimal value could be obtained by cross-validation, and {\small $\vk(\vx^\star)$} is the kernel vector of the sketched data samples and the given test one.

\begin{figure}[h]
  \includegraphics[width=\textwidth]{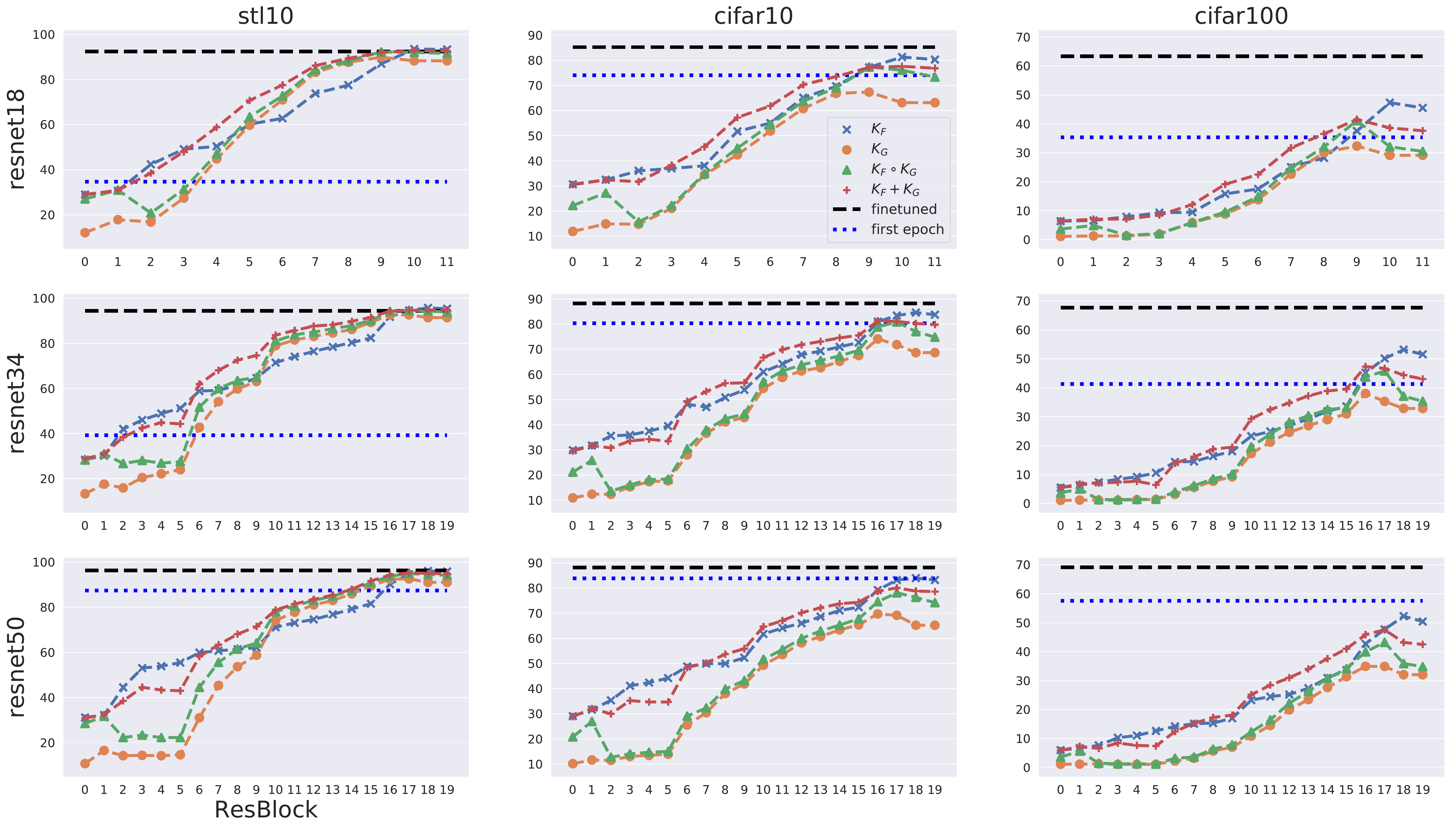}
  \caption{Accuracy on three tasks given the sketched kernels at individual residual blocks. X-axis indicates the index of each residual blocks, and the index goes larger when the layer gets closer to the output layer. Y-axis indicates the chosen model pretrained on ImageNet dataset. The flat dashed line indicates the performance of finetuning the last layer until convergence, and the flat dotted one indicates that of finetuning the last layer for only one epoch.}
  \label{kernelridgeregression}
\end{figure}

Figure \ref{kernelridgeregression} presents the performance of our method with sketching in the framework of kernel ridge regression for classification tasks. One can observe a clear increasing trend of the performance as the chosen layer gets closer to the output layer.

We also finetuned the last layer of pretrained models on individual tasks until convergence for comparison, and there is still a gap between the performance of finetuned models and that of our method. However, since our method with sketching only requires forwarding the dataset only once and then make predictions, we also plotted out the performance of models after finetuning for only one epoch as flat dotted lines. In this comparison, our method provides better performance overall.

There are questions to be asked here. \citet{Avron2016SharperBF} states that, in problem settings of ridge regression, sketching both the data matrix and the target matrix gives a suboptimal solution compared to the analytical solution. When a large number of data samples and a deep neural network is given, it becomes difficult to empirically check how accurate {\small $\vt^\star$} is. Another question is the concern on the effect sketching brings in when casting classification as ridge regression. Both are very interesting for future study.

\end{document}